\documentclass[conference]{IEEEtran}
\IEEEoverridecommandlockouts

\usepackage{cite}
\usepackage{amsmath,amssymb,amsfonts}
\usepackage{graphicx}
\usepackage{textcomp}
\usepackage{xcolor}
\usepackage{url}
\usepackage{listings}
\usepackage{booktabs}
\usepackage[hidelinks]{hyperref}
\usepackage{tikz}
\usetikzlibrary{shapes.geometric, positioning, calc, patterns, arrows.meta}
\usepackage{subcaption}
\usepackage[T1]{fontenc}
\usepackage[utf8]{inputenc}
\usepackage{cuted}
\usepackage{caption}

\urldef\armvideourl\url{https://youtube.com/shorts/Jsz-jFdT3wY}
\urldef\videotwourl\url{https://youtube.com/shorts/Vg3pZLFrEdM}
\urldef\videothreeurl\url{https://youtube.com/shorts/ZCEVujG9srs}
\urldef\repourl\url{https://github.com/Lasan-Perera/6-dof-arm-neuralnexus}
\urldef\repotagurl\url{https://github.com/Lasan-Perera/6-dof-arm-neuralnexus/releases/tag/v1.0}
\urldef\firmwarerepourl\url{https://github.com/Lasan-Perera/neuralnexusarm-codebase}

\newcommand{\blfootnote}[1]{%
  \begingroup
  \renewcommand\thefootnote{}\footnote{#1}%
  \addtocounter{footnote}{-1}%
  \endgroup
}

\begin{document}

\title{Reproducible Vision-Guided 6-DOF\\
Robotic Manipulator with a Mixed Stepper-Driver\\
Architecture and Browser-Native Control}

\author{
\IEEEauthorblockN{%
Lasan Perera\textsuperscript{1,*,$\dagger$},
Deneth Priyadarshana\textsuperscript{1,$\dagger$},
Dulana Pitiwaduge\textsuperscript{1,$\dagger$},
Isitha Dinujaya\textsuperscript{1,$\dagger$}, and
Mokshan Colambage\textsuperscript{1,$\dagger$}}
\IEEEauthorblockA{%
\textsuperscript{1}{Department of Electronic and Telecommunication Engineering}, {University of Moratuwa}, {Moratuwa}, Sri Lanka\\[3pt]
\footnotesize
\textsuperscript{$\dagger$}Equal contribution.\quad  
\textsuperscript{*}Corresponding author: {pererawals.23@uom.lk}, {lasanperera.lsp@gmail.com}\\[3pt]
GitHub Repository: \repourl}
}

\maketitle
\blfootnote{Video 1 (arm overview): \armvideourl. Additional videos are listed in the Availability section.}

\begin{strip}
\centering
\includegraphics[width=\textwidth]{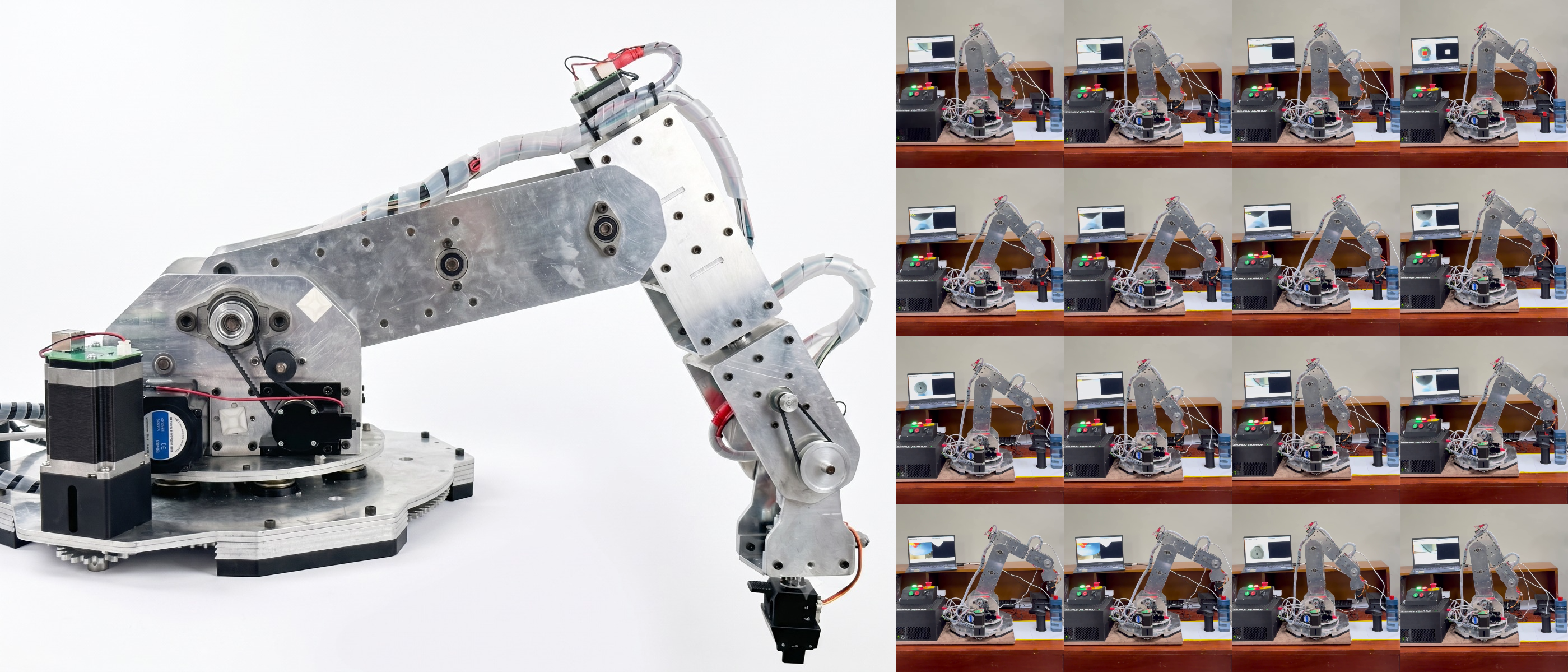}
\captionof{figure}{The NeuralNexus Arm. \textbf{Left:} the assembled 6-DOF
manipulator, showing the base rotation stage, shoulder and elbow driveline, and
wrist assembly. \textbf{Right:} selected frames from a pick-and-place sequence
executed via the browser-native control interface, with real-time camera
feedback (laptop display) and the emergency-stop/status panel visible alongside
the controller.}
\label{fig:teaser}
\end{strip}

\begin{abstract}
We present the NeuralNexus Arm, an open, low-cost 6-DOF robotic manipulator built by an undergraduate engineering team, together with the design decisions and debugging experience needed to reproduce it. The arm is driven by a single STM32H743 microcontroller on a custom printed circuit board (PCB) and combines two stepper-driver strategies on one controller: push--pull 3.3\,V step/direction outputs for onboard TMC2209 drivers on the three wrist joints, and open-drain outputs for external CL57T and DM542 drivers on the three high-torque proximal joints. We describe the mechanical design, mixed-driver electronics, interrupt-driven firmware, a MATLAB/Simscape-based inverse-kinematics pipeline, a browser-native control interface using the Web Serial API, and a lightweight vision pipeline for object localisation and autonomous pick-and-place tasks. We also document non-obvious hardware and firmware failure modes encountered during the transition from a development board to the custom PCB as reproducibility guidance. All design files and firmware are released openly. The platform actuates all six axes under coordinated control at a 2\,kHz update rate and executes both manual and pre-recorded motions from the browser interface.
\end{abstract}


\section{Introduction}
Low-cost 6-DOF robotic arms generally fall into two categories: hobby-grade
manipulators built from RC servos, which are inexpensive but limited in torque,
repeatability, and payload; and industrial or research-grade arms, which are
accurate but costly and difficult to reproduce or modify. There is comparatively
little openly documented middle ground: a stepper-driven arm that a small team
can build, debug, and control without proprietary tooling.

This paper describes such a platform. The NeuralNexus Arm uses stepper actuators
across all six joints and a single STM32H743 controller, and it deliberately
mixes two driver technologies on the same board to match each joint's torque
requirement. Rather than presenting only the finished device, we also document
the engineering path---including a development-board-to-custom-PCB migration and
the cascade of hardware and firmware issues it exposed---so that the design is
genuinely reproducible.

\noindent\textbf{Contributions.} This work contributes:
\begin{itemize}
  \item An open, low-cost 6-DOF manipulator design with a full description of
        the mechanics, electronics, and firmware.
  \item A \emph{mixed stepper-driver architecture} in which onboard TMC2209
        drivers (push--pull, 3.3\,V logic) and external closed-loop drivers
        (open-drain, common-anode wiring) coexist on one microcontroller, with
        a discussion of the polarity and logic subtleties this creates.
  \item A browser-native control interface using the Web Serial API, removing
      any host-side software dependency for jogging and pre-recorded motion.
  \item A documented set of reproducibility lessons---failure modes and their
        root causes---from the custom-PCB bring-up.
\end{itemize}

\section{Related Work}
Open-source robotic manipulators have made robotics research and education more accessible by providing affordable platforms with publicly available mechanical designs, electronics, and software. Notable examples include \textit{HELENE}, a 6-DOF robotic arm featuring closed-loop stepper motors and ROS integration for research and educational use~\cite{Herbst2025HELENE}, and \textit{iArm}, a low-cost 6-DOF educational manipulator built on an open ROS-based software framework~\cite{Zeng2022iArm}. Other educational platforms, such as EduSCARA~\cite{Clark2026EduSCARA} and the PlatROB modular mobile robotics platform~\cite{Balbuena2026PlatROB}, emphasize modular hardware, custom embedded controllers, and low-cost construction. Table~\ref{tab:comparison} compares these platforms with the NeuralNexus Arm.
\begin{table*}[t]
\centering
\caption{Comparison of NeuralNexus with representative open-source robotic manipulators.}
\label{tab:comparison}
\renewcommand{\arraystretch}{1.3}
\footnotesize
\begin{tabular}{@{}p{1.6cm} p{1.1cm} p{0.5cm} p{2.6cm} p{2.2cm} p{6.3cm}@{}}
\hline
\textbf{Robot} &
\textbf{Cost (\$)} &
\textbf{DOF} &
\textbf{Controller} &
\textbf{Motor} &
\textbf{Key Difference from NeuralNexus} \\
\hline
HELENE &
$<1160$ &
6 &
ESP32 + ROS Host &
Stepper &
Single closed-loop stepper driver family throughout (no mixed architecture); ROS-based control, not browser-native; the only competitor with a published ISO~9283 repeatability validation. \\

iArm &
$\approx250$ &
6 &
Raspberry Pi 4 + ROS &
Serial Bus Servo &
Servo-actuated, not stepper-driven; far lower torque/payload class; ROS-based host software rather than a zero-install browser interface. \\

EduSCARA &
$\approx150$ &
4 &
STM32 Nucleo + Python API &
Hobby Servo &
Only 4-DOF SCARA kinematics (RRPR), not a full 6-DOF articulated arm; potentiometer feedback rather than closed-loop stepper/encoder joints. \\

AR4 (MK3/MK4) &
$\approx2000$ &
6 &
Teensy 4.1 + Arduino Nano/Mega &
Open-Loop Stepper &
Uses one driver family only (DM542T/DM320T, same open-loop class at different current ratings) -- not a mixed architecture; encoders exist but closed-loop correction is not implemented in firmware; Arduino/ROS control, no browser-native UI; non-commercial license restricts redistribution/resale. \\

PAROL6 &
$\gtrsim350$ &
6 &
Custom STM32 Board + Python Commander GUI &
Stepper (Closed-Loop Upgradable) &
Single driver family on its custom board (closed-loop is an optional FOC-driver upgrade, not a factory heterogeneous design); control is via a desktop Python app, not a browser; closest analog to our custom-PCB approach but without driver-family mixing. \\

Forte &
$<215$ &
6 &
Off-the-Shelf Drivers, No Custom Board &
Stepper (Capstan/Belt Drive) &
No custom controller PCB at all -- uses off-the-shelf drivers with a host PC; achieves low cost via a capstan-cable/belt drivetrain rather than gearbox+belt reductions; open-source release status unconfirmed. \\

NeuralNexus &
$\approx1512$ &
6 &
STM32H743 + Web Serial UI &
Closed-Loop Stepper &
--- \\
\hline
\end{tabular}
\end{table*}

In contrast, the proposed \textit{NeuralNexus Arm} focuses on controller-level design and reproducibility rather than mechanical novelty. The system combines a mixed stepper-driver architecture with a custom STM32H743-based controller, browser-native control through the Web Serial API, and comprehensive documentation covering PCB design, firmware, system integration, and bring-up procedures to facilitate straightforward replication by researchers and student engineering teams.

\section{Mechanical Design}
\begin{figure}[t]
  \centering
  \includegraphics[width=\columnwidth]{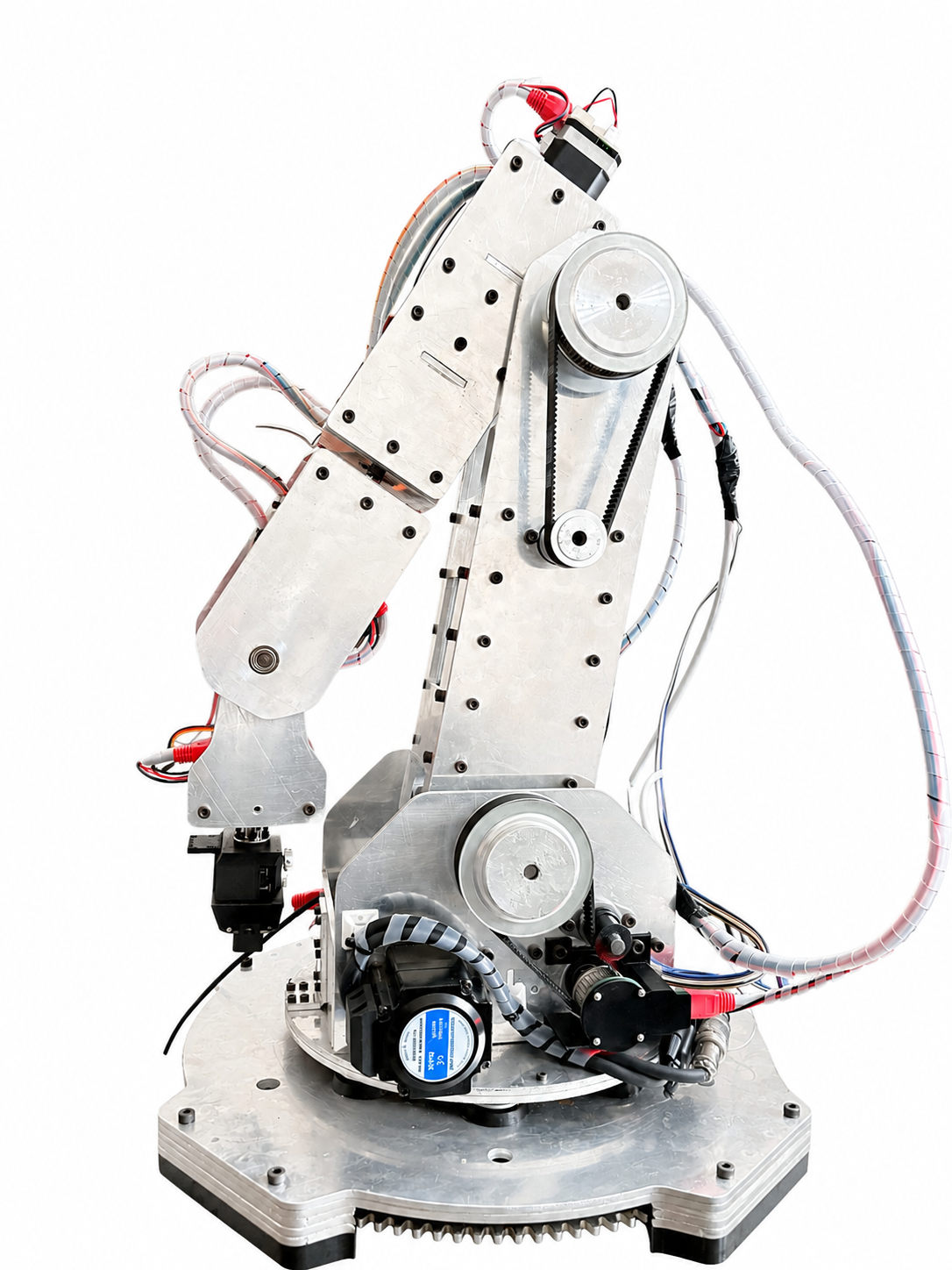}
  \caption{The assembled NeuralNexus Arm.}
  \label{fig:arm}
\end{figure}

The NeuralNexus Arm (Fig.~\ref{fig:arm}) is a 6-DOF serial robotic
manipulator designed with a hierarchical actuation architecture to satisfy the
varying torque and speed requirements of each joint\cite{Vyas2022Economic,ScheinmanMechanism}. High-torque proximal
joints employ NEMA~23 and NEMA~24 stepper motors combined with planetary
gearboxes and timing-belt reductions, while the distal wrist joints utilize
compact NEMA~17 stepper motors to reduce the moving inertia of the manipulator\cite{Herbst2025HELENE,Chebly2025Forte,Kim2025ARMADA}.
This approach improves dynamic performance without compromising the payload
capacity of the arm\cite{Kim2025ARMADA}.

Power transmission is primarily achieved using HTD3M timing belts with a belt
width of 15\,mm for the high-load joints, providing reliable torque transmission with minimal slip and backlash\cite{TimingBeltHandbook,Shigley}. The wrist joints employ lighter GT2 timing
belts where lower transmitted torque and compact dimensions are advantageous\cite{TimingBeltHandbook}.
The complete actuation system for each joint is described in the following
subsections.

\subsection{Base Rotation (J1)}                                                                                                                               

\begin{figure}[htbp]
    \centering
\begin{tikzpicture}[
    >=Stealth,
    node distance=0.65cm,
    block/.style={
        draw=navy!80!black,
        fill=blue!5,
        thick,
        rounded corners=3pt,
        minimum width=2.1cm,
        minimum height=1.0cm,
        align=center,
        font=\sffamily\scriptsize
    }
]

    \definecolor{navy}{RGB}{20, 45, 85}

    \node [block] (motor) {
        \textbf{NEMA 23 Stepper}\\[1pt]
        Motor
    };

    \node [block, right=of motor] (gearbox) {
        \textbf{1:7 Spur Gear}\\[1pt]
        Reduction
    };

    \node [right=0.7cm of gearbox, font=\sffamily\bfseries\scriptsize, text=navy] (j1) {J1};

    \draw [->, thick, navy] (motor) -- (gearbox);
    \draw [->, thick, navy] (gearbox) -- (j1);

    \node [above=0.08cm of motor, font=\sffamily\bfseries\tiny, text=gray!60!black] {INPUT};
    \node [above=0.08cm of gearbox, font=\sffamily\bfseries\tiny, text=gray!60!black] {STAGE 1};
    \node [above=0.08cm of j1, font=\sffamily\bfseries\tiny, text=gray!60!black] {OUTPUT};

\end{tikzpicture}
\label{fig:driveline_schematic_nema23}
\end{figure}

The base joint supports the entire manipulator and provides rotation about the
vertical axis. To generate the high output torque required for this motion, a
NEMA~23 stepper motor is coupled to the base through a 1:7 spur gear reduction.
The large reduction ratio increases the available output torque while improving
angular positioning resolution and reducing the torque demand on the motor \cite{Shigley,ScheinmanMechanism}.

\subsection{Shoulder Joint (J2)}

\begin{figure}[htbp]
    \centering
\begin{tikzpicture}[
    >=Stealth,
    node distance=0.65cm,
    block/.style={
        draw=navy!80!black,
        fill=blue!5,
        thick,
        rounded corners=3pt,
        minimum width=2.1cm,
        minimum height=1.0cm,
        align=center,
        font=\sffamily\scriptsize
    }
]

    \definecolor{navy}{RGB}{20, 45, 85}

    \node [block] (motor) {
        \textbf{NEMA 24 Stepper}\\[1pt]
        Motor
    };

    \node [block, right=of motor] (gearbox) {
        \textbf{1:10 Planetary}\\[1pt]
        Gearbox
    };

    \node [block, right=of gearbox] (pulley) {
        \textbf{30T:75T HTD3M}\\[1pt]
        Belt Reduction
    };

    \node [right=0.7cm of pulley, font=\sffamily\bfseries\scriptsize, text=navy] (j1) {J2};

    \draw [->, thick, navy] (motor) -- (gearbox);
    \draw [->, thick, navy] (gearbox) -- (pulley);
    \draw [->, thick, navy] (pulley) -- (j1);

    \node [above=0.08cm of motor, font=\sffamily\bfseries\tiny, text=gray!60!black] {INPUT};
    \node [above=0.08cm of gearbox, font=\sffamily\bfseries\tiny, text=gray!60!black] {STAGE 1};
    \node [above=0.08cm of pulley, font=\sffamily\bfseries\tiny, text=gray!60!black] {STAGE 2};
    \node [above=0.08cm of j1, font=\sffamily\bfseries\tiny, text=gray!60!black] {OUTPUT};

\end{tikzpicture}
\label{fig:driveline_j2}
\end{figure}

The shoulder joint carries a significant portion of the manipulator mass and is
therefore designed for high torque\cite{Herbst2025HELENE,Jazar2022}. The joint is actuated using a NEMA~24
stepper motor coupled to a 1:10 planetary gearbox\cite{Garcia2020Gearboxes,Hrdlicka2022Gearbox}. The gearbox output is
further transmitted through an HTD3M synchronous timing-belt drive using a
30-tooth driving pulley and a 75-tooth driven pulley, providing an additional
1:2.5 reduction\cite{Shigley,TimingBeltHandbook}. The HTD3M 15\,mm timing belt was selected for its high torque
capacity and reliable power transmission\cite{TimingBeltHandbook}.

\subsection{Elbow Joint (J3)}

\begin{figure}[htbp]
    \centering
\begin{tikzpicture}[
    >=Stealth,
    node distance=0.65cm and 0.8cm, 
    block/.style={
        draw=navy!80!black,
        fill=blue!5,
        thick,
        rounded corners=3pt,
        minimum width=2.2cm,
        minimum height=0.95cm,
        align=center,
        font=\sffamily\scriptsize
    }
]

    \definecolor{navy}{RGB}{20, 45, 85}

    \node [block] (motor) {
        \textbf{NEMA 24 Stepper}\\[1pt]
        Motor
    };

    \node [block, right=of motor] (gearbox) {
        \textbf{1:10 Planetary}\\[1pt]
        Gearbox
    };

    \node [block, below=0.65cm of gearbox] (transfer) {
        \textbf{30T:30T HTD3M}\\[1pt]
        Pulley Transfer
    };

    \node [block, left=of transfer] (pulley) {
        \textbf{30T:75T HTD3M}\\[1pt]
        Pulley Reduction
    };

    \node [left=0.6cm of pulley, font=\sffamily\bfseries\scriptsize, text=navy] (j1) {J3};

    \draw [->, thick, navy] (motor) -- (gearbox);
    \draw [->, thick, navy] (gearbox) -- (transfer);
    \draw [->, thick, navy] (transfer) -- (pulley);
    \draw [->, thick, navy] (pulley) -- (j1);

    \node [above=0.08cm of motor, font=\sffamily\bfseries\tiny, text=gray!60!black] {INPUT};
    \node [above=0.08cm of gearbox, font=\sffamily\bfseries\tiny, text=gray!60!black] {STAGE 1};
    \node [below=0.08cm of transfer, font=\sffamily\bfseries\tiny, text=gray!60!black] {STAGE 2};
    \node [below=0.08cm of pulley, font=\sffamily\bfseries\tiny, text=gray!60!black] {STAGE 3};
    \node [below=0.08cm of j1, font=\sffamily\bfseries\tiny, text=gray!60!black] {OUTPUT};

\end{tikzpicture}
\label{fig:driveline_j3}
\end{figure}

The elbow joint extends the manipulator workspace while supporting the distal
links and payload\cite{Jazar2022,ScheinmanMechanism}. The actuation system consists of a NEMA~24 stepper motor
connected to a 1:10 planetary gearbox\cite{Garcia2020Gearboxes,Hrdlicka2022Gearbox}. Motion is transferred through two
intermediate HTD3M timing-belt stages with identical 30-tooth pulleys,
followed by a final HTD3M timing-belt reduction using a 30-tooth driving pulley
and a 75-tooth driven pulley\cite{TimingBeltHandbook}. The intermediate belt stages relocate the motor
to a mechanically convenient position while the final stage provides the
required torque multiplication\cite{TimingBeltHandbook,Shigley}.

\subsection{Wrist Pitch (J4)}

\begin{figure}[htbp]
    \centering
\begin{tikzpicture}[
    >=Stealth,
    node distance=0.65cm,
    block/.style={
        draw=navy!80!black,
        fill=blue!5,
        thick,
        rounded corners=3pt,
        minimum width=2.1cm,
        minimum height=1.0cm,
        align=center,
        font=\sffamily\scriptsize
    }
]

\definecolor{navy}{RGB}{20,45,85}

\node [block] (motor) {
    \textbf{NEMA 17 Stepper}\\[1pt]
    Motor
};

\node [block, right=of motor] (gearbox) {
    \textbf{1:4 Planetary}\\[1pt]
    Gearbox
};

\node [right=0.7cm of gearbox,
       font=\sffamily\bfseries\scriptsize,
       text=navy] (j4) {J4};

\draw [->, thick, navy] (motor) -- (gearbox);
\draw [->, thick, navy] (gearbox) -- (j4);

\node [above=0.08cm of motor,
       font=\sffamily\bfseries\tiny,
       text=gray!60!black] {INPUT};

\node [above=0.08cm of gearbox,
       font=\sffamily\bfseries\tiny,
       text=gray!60!black] {STAGE 1};

\node [above=0.08cm of j4,
       font=\sffamily\bfseries\tiny,
       text=gray!60!black] {OUTPUT};

\end{tikzpicture}
\label{fig:j3_driveline}
\end{figure}

The first wrist joint is actuated using a compact NEMA~17 stepper motor
combined with a 1:4 planetary gearbox\cite{Garcia2020Gearboxes,Hrdlicka2022Gearbox}. The integrated gearbox provides
sufficient output torque while maintaining a compact mechanical design suitable
for the wrist assembly\cite{Garcia2020Gearboxes,Hrdlicka2022Gearbox}.

\subsection{Wrist Roll (J5)}

\begin{figure}[htbp]
    \centering
\begin{tikzpicture}[
    >=Stealth,
    node distance=0.65cm,
    block/.style={
        draw=navy!80!black,
        fill=blue!5,
        thick,
        rounded corners=3pt,
        minimum width=2.3cm,
        minimum height=1.0cm,
        align=center,
        font=\sffamily\scriptsize
    }
]

\definecolor{navy}{RGB}{20,45,85}

\node [block] (motor) {
    \textbf{NEMA 17 Stepper}\\[1pt]
    Motor
};

\node [block, right=of motor] (belt) {
    \textbf{20T:60T GT2}\\[1pt]
    Belt Reduction
};

\node [right=0.7cm of belt,
       font=\sffamily\bfseries\scriptsize,
       text=navy] (j5) {J5};

\draw [->, thick, navy] (motor) -- (belt);
\draw [->, thick, navy] (belt) -- (j5);

\node [above=0.08cm of motor,
       font=\sffamily\bfseries\tiny,
       text=gray!60!black] {INPUT};

\node [above=0.08cm of belt,
       font=\sffamily\bfseries\tiny,
       text=gray!60!black] {STAGE 1};

\node [above=0.08cm of j5,
       font=\sffamily\bfseries\tiny,
       text=gray!60!black] {OUTPUT};

\end{tikzpicture}
\label{fig:j4_driveline}
\end{figure}

The wrist roll joint is driven by a NEMA~17 stepper motor through a GT2 timing
belt transmission using a 20-tooth driving pulley and a 60-tooth driven pulley,
providing a 1:3 reduction ratio\cite{TimingBeltHandbook,Shigley}. A 6\,mm wide GT2 timing belt was selected to
achieve compact packaging while providing adequate torque transmission for the
lightweight wrist mechanism\cite{TimingBeltHandbook,Kim2025ARMADA}.

\subsection{Gripper Rotation (J6)}

\begin{figure}[htbp]
    \centering
\begin{tikzpicture}[
    >=Stealth,
    node distance=0.65cm,
    block/.style={
        draw=navy!80!black,
        fill=blue!5,
        thick,
        rounded corners=3pt,
        minimum width=2.3cm,
        minimum height=1.0cm,
        align=center,
        font=\sffamily\scriptsize
    }
]

\definecolor{navy}{RGB}{20,45,85}

\node [block] (motor) {
    \textbf{NEMA 17 Pancake}\\[1pt]
    Stepper Motor
};

\node [block, right=of motor] (drive) {
    \textbf{Direct Drive}\\[1pt]
    No Reduction
};

\node [right=0.7cm of drive,
       font=\sffamily\bfseries\scriptsize,
       text=navy] (j6) {J6};

\draw [->, thick, navy] (motor) -- (drive);
\draw [->, thick, navy] (drive) -- (j6);

\node [above=0.08cm of motor,
       font=\sffamily\bfseries\tiny,
       text=gray!60!black] {INPUT};

\node [above=0.08cm of drive,
       font=\sffamily\bfseries\tiny,
       text=gray!60!black] {STAGE 1};

\node [above=0.08cm of j6,
       font=\sffamily\bfseries\tiny,
       text=gray!60!black] {OUTPUT};

\end{tikzpicture}
\label{fig:j6_driveline}
\end{figure}

The rotational motion of the end effector is provided by a pancake NEMA~17
stepper motor mounted directly within the wrist assembly. The compact axial
length of the pancake motor minimizes the overall size and inertia of the wrist
while providing sufficient torque for orienting the gripper\cite{Chebly2025Forte,Kim2025ARMADA}.

\subsection{Gripper}

Object grasping is achieved using a parallel-jaw gripper actuated by a digital
servo motor. The servo provides closed-loop position control, enabling precise
opening and closing of the gripper while simplifying the mechanical design\cite{ScheinmanMechanism}.
Separating the gripping mechanism from the rotational wrist joint allows
independent control of end-effector orientation and grasping force.

\begin{figure}[t]
  \centering
  \includegraphics[width=\columnwidth]{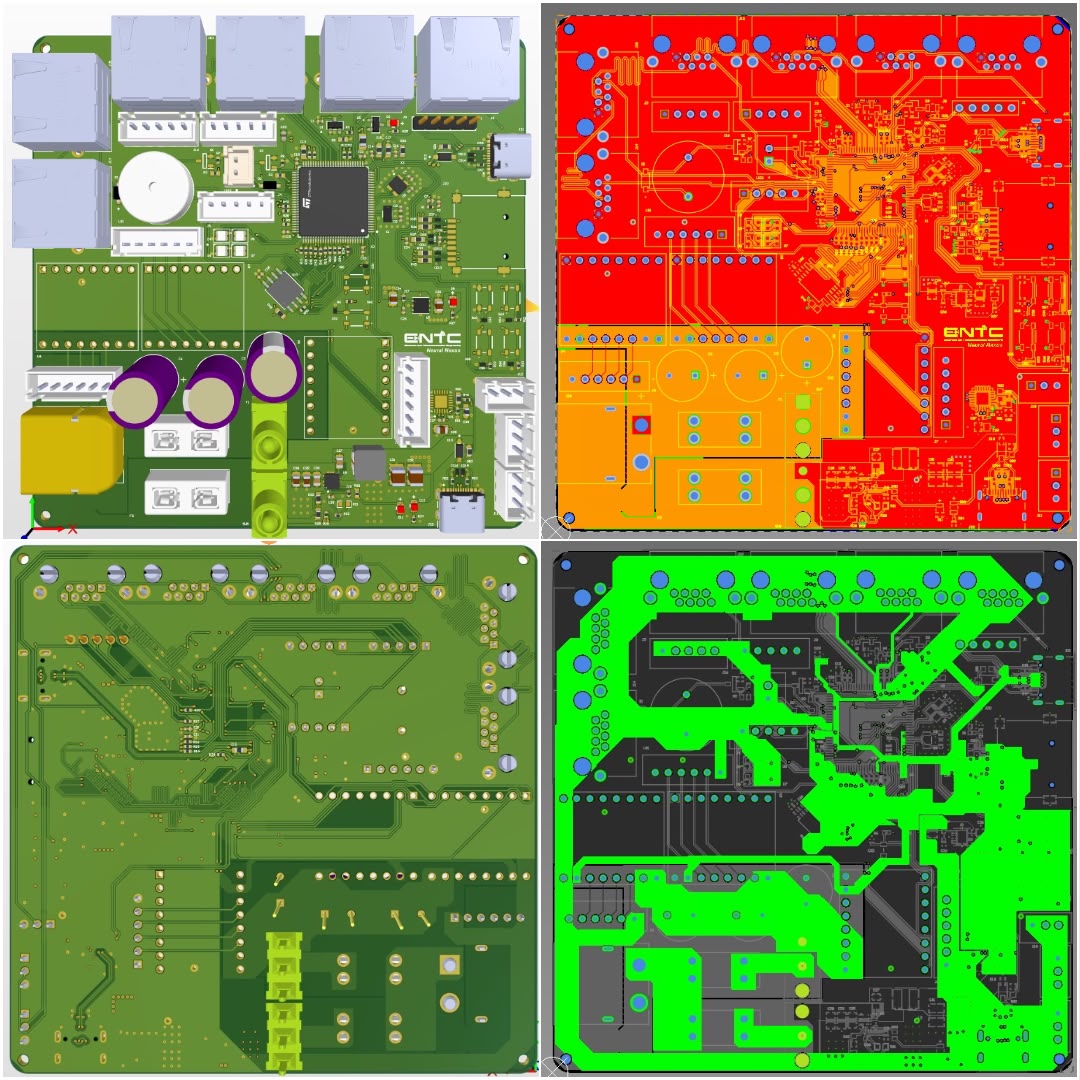}
  \caption{Custom STM32H743 controller PCB, showing the onboard TMC2209 drivers
  for the wrist joints and the terminals to the external closed-loop drivers.}
  \label{fig:pcb}
\end{figure}
%
\section{Electronics, Controller \& Firmware Architecture}
\label{sec:firmware}
The controller is a custom PCB (Fig.~\ref{fig:pcb}) built around the STM32H743VIT6, a 480\,MHz Arm Cortex-M7 microcontroller\cite{STM32H743Datasheet,STM32H743Reference}. A single microcontroller generates the step and direction
signals for all six joints, reads the joint encoders, and communicates with the
host over serial. The onboard wrist-motor drivers, the interface to the external
proximal-joint drivers, the power conditioning, and the motor and signal
connectors are integrated on one board. The central design decision is that the
six joints are \emph{not} driven uniformly: the torque demands of the proximal
and wrist joints differ by roughly an order of magnitude, so two different driver
technologies are combined on the same controller.

\subsection{Actuator and Driver Selection}
The three proximal joints support the mass of the entire arm and its payload and
therefore use high-torque motors with \emph{external closed-loop} drivers: a
DM542 driving a NEMA-23 motor at the base (J1), and CL57T closed-loop drivers on
the NEMA-24 shoulder and elbow motors (J2, J3)\cite{DM542Manual,CL57TManual}. Closed-loop drive was chosen for
these axes because a step lost under gravitational load is both likely and
consequential; the driver's internal encoder loop corrects for it and reports
stall\cite{CL57TManual,MicrochipAN1307}. The three wrist joints (J4--J6) carry much lighter loads and use NEMA-17
motors driven by \emph{onboard} TMC2209 drivers, which are compact, quiet, and
adequate for the wrist\cite{TMC2209Datasheet}. Table~\ref{tab:drivers} summarises the mapping.

\begin{table}[t]
  \caption{Per-joint actuator and driver mapping.}
  \label{tab:drivers}
  \centering
  \begin{tabular}{@{}llll@{}}
    \toprule
    Joint & Motor class & Driver & GPIO mode \\
    \midrule
    J1 (base)     & NEMA 23 & DM542 & Open-drain \\
    J2 (shoulder) & NEMA 24 & CL57T & Open-drain \\
    J3 (elbow)    & NEMA 24 & CL57T & Open-drain \\
    J4 (wrist 1)  & NEMA 17 & TMC2209 (onboard) & Push--pull \\
    J5 (wrist 2)  & NEMA 17 & TMC2209 (onboard) & Push--pull \\
    J6 (wrist 3)  & NEMA 17 & TMC2209 (onboard) & Push--pull \\
    \bottomrule
  \end{tabular}
\end{table}

\subsection{Mixed Driver Interfacing}
\label{sec:mixed_driver_interfacing}
The two driver families present different electrical interfaces to the
microcontroller, and the PCB accommodates both.

The onboard TMC2209 drivers accept 3.3\,V logic-level STEP and DIR inputs, so the
corresponding STM32 pins are configured as ordinary \emph{push--pull} outputs and
routed directly to the drivers\cite{TMC2209Datasheet,STM32H743Reference}.

The external CL57T and DM542 drivers instead expose optically isolated inputs
(STEP\texttt{+/-}, DIR\texttt{+/-}, EN\texttt{+/-}) intended for 5\,V signalling\cite{CL57TManual,DM542Manual}.
They are wired in a \emph{common-anode} configuration: the \texttt{+} terminals
are tied to a shared \texttt{+5\,V} rail and the microcontroller sinks current
through the \texttt{-} terminals\cite{CL57TManual,DM542Manual}. Because the STM32 cannot source 5\,V, the pins
driving these inputs are configured as \emph{open-drain} outputs: each pin either
sinks current---driving the input optocoupler from the 5\,V rail---or floats,
leaving it dark\cite{STM32H743Reference}. This allows a 3.3\,V microcontroller to drive nominally 5\,V
opto-isolated inputs without external level shifters.

\subsection{Enable-Signal Polarity}
Enable polarity is not uniform across the arm, and the firmware handles the two
driver families with separate routines (\texttt{Stepper\_SetEnableM1M2M3} and
\texttt{Stepper\_SetEnableM4M5M6}). Note that the firmware motor-channel indices
M1--M6 do not follow joint order; they are mapped to joints J1--J6 through the
\texttt{jointToMotor} table, so the onboard channels M1--M3 correspond to the
wrist joints J4--J6 and the external channels M4--M6 to the proximal joints
J1--J3.

\begin{figure}[t]
\centering
\resizebox{\columnwidth}{!}{%
\begin{tikzpicture}[
  font=\small,
  >={Stealth[length=1.6mm]},
  box/.style={draw, rounded corners=1.5pt, align=center, inner sep=3pt, minimum height=7mm},
  mcu/.style={draw, thick, rounded corners=1.5pt, align=center, fill=black!6,
              minimum width=44mm, minimum height=10mm},
  motor/.style={draw, align=center, inner sep=3pt, minimum height=7mm,
                minimum width=32mm, fill=black!3},
  sig/.style={-{Stealth[length=1.6mm]}, thick},
  fb/.style={-{Stealth[length=1.6mm]}, thick, dashed},
  clab/.style={font=\scriptsize, align=center},
]
\node[box, minimum width=40mm] (host) {Host PC --- Web Serial (USB CDC)};
\node[mcu, below=5mm of host] (mcu) {\textbf{STM32H743VITx}\\[1pt]\footnotesize custom controller PCB\\[1pt]\footnotesize 2\,kHz step generator (TIM6)};

\node[box, below=13mm of mcu, xshift=-24mm, minimum width=36mm] (tmc) {$3\times$ TMC2209\\\footnotesize onboard, wrist};
\node[box, below=13mm of mcu, xshift=24mm, minimum width=40mm] (ext) {$2\times$ CL57T $+$ $1\times$ DM542\\\footnotesize external, closed-loop drivers};

\node[motor, below=5mm of tmc] (wrist) {NEMA-17 $\times 3$\\\footnotesize wrist J4--J6};
\node[motor, below=5mm of ext, minimum width=38mm] (prox) {NEMA-23/24 $\times 3$\\\footnotesize proximal J1--J3};

\draw[sig] (host) -- (mcu);
\draw[sig] (mcu.south) -- (tmc.north)
   node[midway, clab, fill=white, inner sep=1.5pt]{push--pull\\3.3\,V};
\draw[sig] (mcu.south) -- (ext.north)
   node[midway, clab, fill=white, inner sep=1.5pt]{open-drain\\5\,V opto};
\draw[sig] (tmc) -- (wrist);
\draw[sig] (ext) -- (prox);

\draw[fb] (wrist.west) -- ++(-6mm,0) |- (mcu.west);
\draw[fb] (prox.east) -- ++(6mm,0) |- (mcu.east)
   node[pos=0.25, clab, rotate=90, below]{enc.\ (SPI)};

\node[box, below=6mm of wrist, xshift=24mm, minimum width=84mm, fill=black!2]
  (pwr) {\footnotesize motor rail $\rightarrow$ drivers \; $|$ \; 3.3\,V LDO $\rightarrow$ logic};
\end{tikzpicture}%
}
\caption{Controller architecture. A single STM32H743 drives two stepper-driver
families over separate electrical interfaces: push--pull 3.3\,V logic to the
onboard TMC2209 wrist drivers, and open-drain, common-anode 5\,V signalling to the
opto-isolated external closed-loop drivers on the proximal joints. Per-joint
AS5047P magnetic encoders return over SPI\cite{STM32H743Reference,TMC2209Datasheet,CL57TManual,DM542Manual,AS5047PDatasheet}.}
\label{fig:arch}
\end{figure}
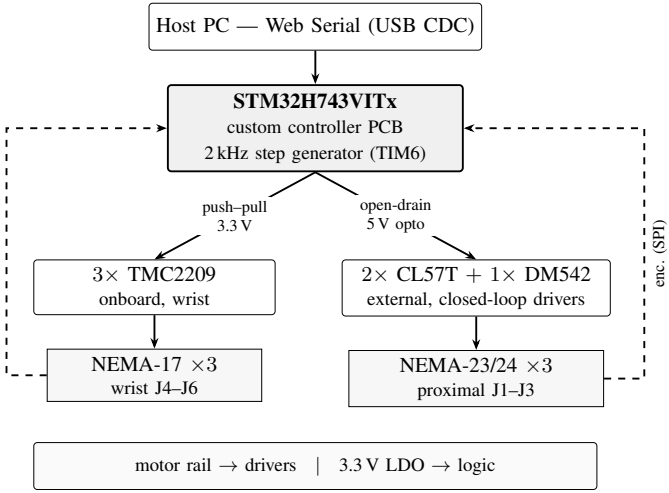

The onboard TMC2209 drivers (M1--M3) use a conventional \emph{active-low,
push--pull} enable: the GPIO is driven low to enable the driver and high to
disable it\cite{TMC2209Datasheet}. The external drivers (M4--M6) use the open-drain, common-anode
interface described above, and---critically---the two external models do not
share the same enable convention. The CL57T closed-loop drivers (shoulder and
elbow) are enabled by a \emph{low} input, in which the microcontroller sinks
current and illuminates the enable optocoupler\cite{CL57TManual}. The DM542 at the base is the
opposite: it is enabled when the open-drain line is \emph{released high} and the
optocoupler is left dark\cite{DM542Manual}. This asymmetry follows from the differing ENA-input
conventions of the two driver models rather than from the wiring itself, so the
enable line for each external axis is verified individually on hardware. Because
the resulting polarity is neither uniform nor intuitive, it is documented
explicitly in the firmware and must be preserved through any refactor.

\subsection{Microstepping and Step Timing}

The overall controller architecture is shown in Fig.~\ref{fig:arch}.

Microstep resolution and phase current are configured in hardware---the external
CL57T and DM542 drivers through their DIP switches, and the onboard TMC2209
drivers through their configuration (MS1/MS2) pins---and are not changed at run
time\cite{CL57TManual,DM542Manual,TMC2209Datasheet}. Instead the firmware stores, per joint, the number of step pulses
corresponding to one full joint revolution, which combines the driver's microstep
resolution with the joint's gear reduction\cite{MicrochipAN907}. The wrist joint (J6) is driven
directly, so its pulses per revolution equal the microstep setting times the
motor's 200 full steps per revolution; all other joints (J1--J5) additionally
include their gear ratio\cite{MicrochipAN907}. The currently configured values (pulses per joint
revolution, J1--J6) are 2800, 5000, 5000, 6400, 1600, and 1600. These are being
calibrated per joint with a dedicated routine that commands a known angle and
compares it against the absolute encoder: J6 was found to require 3200 pulses per
revolution rather than the initially assumed 1600.

Step pulses are produced by a single timer interrupt (TIM6) running at 2\,kHz,
i.e.\ a 500\,$\mu$s control tick, consistent with the firmware timebase
($dt=0.5$\,ms)\cite{STM32H743Reference}. Each STEP line is raised on the tick at which a step falls due and
cleared at the start of the following tick, so every pulse is held high for one
full tick ($\approx$500\,$\mu$s). This is far wider than the minimum STEP
pulse-width required by any of the drivers---in particular the opto-isolated CL57T
and DM542 inputs, whose optocouplers need a much wider pulse than the TMC2209---so
a single generator drives both driver families without a separate fast path\cite{TMC2209Datasheet,CL57TManual,DM542Manual}. The
STEP lines of the external axes are open-drain and pulled up through the
common-anode 5\,V rail; the 500\,$\mu$s high time comfortably accommodates their
slower rise. The DIR line for each axis is set at the start of a move and held
constant for its duration, so direction setup time is never a limiting factor\cite{TMC2209Datasheet,CL57TManual,DM542Manual}.
Per-axis step rate is governed by a trapezoidal velocity ramp through a step
interval of $\lfloor 2000/\text{speed} \rfloor$ ticks, with joint speeds currently
capped at 400\,steps/s \cite{MicrochipAN1307}.

\subsection{TMC2209 Configuration}
The onboard TMC2209 drivers default to StealthChop, which is quiet at low speed
but sheds torque as the step rate rises\cite{TMC2209Datasheet}. To retain torque during motion the
SPREAD pin is tied to VIO, selecting SpreadCycle\cite{TMC2209Datasheet}. The drivers are used in
standalone (pin-configured) step/direction mode, and each driver's phase-current
limit is set through its VREF pin together with the on-board 0.11\,$\Omega$ sense
resistors\cite{TMC2209Datasheet}. The wrist uses NEMA-17 motors of two ratings---standard units at
1.5\,A per phase and a lower-profile pancake unit at 0.8\,A per phase---so VREF is
set per driver to match each motor: at its 3.3\,V full-scale maximum for the
1.5\,A axes, and at a correspondingly lower value for the 0.8\,A pancake axis,
since VREF scales the RMS phase current approximately linearly\cite{TMC2209Datasheet}.

\subsection{Power Architecture}
The board separates the motor-supply rail from the logic supply, listed by
motor class in Table~\ref{tab:power}. The stepper motors and their drivers are fed from the main motor rail,
\begin{table}[ht]
  \caption{Motor-supply rails by motor class.}
  \label{tab:power}
  \centering
  \begin{tabular}{@{}lll@{}}
    \toprule
    Motor class & Joint(s) & Supply voltage \\
    \midrule
    NEMA 17 & J4--J6 (wrist)           & 12\,V \\
    NEMA 23 & J1 (base)                & 24\,V \\
    NEMA 24 & J2, J3 (shoulder, elbow) & 48\,V \\
    \midrule
    5\,V (opto)   & J1--J3 driver inputs      & 5\,V (buck from 12\,V) \\
    Logic (MCU) & ---                  & 3.3\,V (LDO) \\
    \bottomrule
  \end{tabular}
\end{table}while the microcontroller and logic circuitry are
powered at 3.3\,V by an on-board ST1L05CPU33R low-dropout regulator\cite{ST1L05Datasheet}. The
opto-isolated $+5$\,V rail used by the external CL57T and DM542 driver inputs
(Section~\ref{sec:mixed_driver_interfacing}) is generated from the 12\,V wrist
rail by a TPS56637RPAR synchronous buck converter\cite{TPS56637Datasheet},
rather than a dedicated supply. Motor and
logic grounds are joined at a single point to limit switching noise on the logic
rail\cite{TIMotorDriverLayout}. During bring-up, the custom PCB also required three GPIO reassignments to
route around damaged pins (PE0$\rightarrow$PE7, PD7$\rightarrow$PD11 via a bodge
wire on the M3\_MS1 pad, and PD3$\rightarrow$PD12); these, and an observed
regulator failure, are discussed as reproducibility guidance in the
lessons-learned section, where we recommend verifying the actual regulator input
voltage before power-up.
\section{Kinematics and Motion Planning}

\subsection{Kinematic Model}

The kinematic model of the NeuralNexus Arm was developed using the MATLAB
Robotics System Toolbox by representing the manipulator as a
\texttt{rigidBodyTree}\cite{MATLABRobotics}. The model consists of six rigid bodies connected by six
revolute joints (J1--J6), with the geometric relationship between consecutive
links defined by fixed homogeneous transformations extracted from the validated
Simscape Multibody model. This representation provides a consistent numerical
model for forward kinematics, inverse kinematics, and trajectory generation
while maintaining agreement with the mechanical design.

Unlike conventional Denavit--Hartenberg (DH) modelling, where coordinate frames
must be manually assigned, the \texttt{rigidBodyTree} directly stores the rigid
transformations between adjacent joint frames. Consequently, the imported model
preserves the original CAD-based geometry without requiring an intermediate DH
parameterisation.

The homogeneous transformation between two consecutive joint frames is
represented as

\begin{equation}
{}^{i-1}\mathbf{T}_{i}=
\begin{bmatrix}
\mathbf{R}_{i} & \mathbf{p}_{i}\\
\mathbf{0}^{T} & 1
\end{bmatrix},
\end{equation}

where $\mathbf{R}_{i}\in SO(3)$  \cite{LynchPark} denotes the rotation matrix and
$\mathbf{p}_{i}\in\mathbb{R}^{3}$ denotes the translation vector between the
parent and child joint frames.

The forward kinematics of the manipulator is obtained by the ordered product of
the individual link transformations\cite{Craig,Spong},

\begin{equation}
{}^{0}\mathbf{T}_{6}
=
\prod_{i=1}^{6}
{}^{i-1}\mathbf{T}_{i},
\end{equation}

which provides the end-effector pose relative to the base coordinate frame.

Table~\ref{tab:kinematics} summarises the fixed translations and joint axes
exported directly from the MATLAB \texttt{rigidBodyTree} model. These
transformations completely define the kinematic chain used by the numerical
inverse kinematics solver.

\begin{table}[!t]
\centering
\caption{Joint frame transformations extracted from the MATLAB
\texttt{rigidBodyTree} model.}
\label{tab:kinematics}

\begin{tabular}{cccc}
\toprule
Joint & Translation (m) & Joint Axis & Type\\
\midrule
J1 & $(0,\;0.017,\;0)$ & $(0,0,-1)$ & Revolute\\
J2 & $(0,\;-0.075,\;-0.138)$ & $(0,0,1)$ & Revolute\\
J3 & $(0.310,\;0,\;-0.024)$ & $(0,0,1)$ & Revolute\\
J4 & $(-0.060,\;0.160,\;-0.051)$ & $(0,0,1)$ & Revolute\\
J5 & $(0.041,\;0,\;-0.1065)$ & $(0,0,1)$ & Revolute\\
J6 & $(0,\;0.110,\;-0.041)$ & $(0,0,1)$ & Revolute\\
\bottomrule
\end{tabular}
\end{table}

Mechanical joint limits were incorporated into the
\texttt{rigidBodyTree} model according to the allowable motion of each joint.
These limits constrain the inverse kinematics optimisation to produce only
physically achievable configurations.


\subsection{Inverse Kinematics}

Desired end-effector poses are converted into joint angles using the MATLAB
Robotics System Toolbox \texttt{inverseKinematics} solver\cite{MATLABRobotics,LynchPark}. Rather than deriving
a closed-form analytical solution, the solver employs a numerical optimisation
procedure that minimises the Cartesian pose error between the desired and
computed end-effector configurations\cite{Nocedal}.

The optimisation is initialised using the current joint configuration,
allowing rapid convergence while maintaining configuration continuity between
successive target poses. Equal weighting was assigned to the translational and
rotational components of the pose error using the weighting vector

\[
\mathbf{w}
=
[1\;1\;1\;1\;1\;1],
\]

thereby giving equal importance to position and orientation during the
optimisation process.

Joint limits defined within the \texttt{rigidBodyTree} are enforced throughout
the optimization, preventing solutions outside the mechanical operating range
of the manipulator\cite{Spong}.

During system validation and software testing, the reference joint
configuration

\[
\boldsymbol{\theta}
=
[\,0,\,-5,\,120,\,0,\,160,\,0\,]^{\circ}
\]

was used as a nominal pose for verifying the forward and inverse kinematic
models.


\subsection{Motion Execution}

\begin{figure}[!t]
\centering
\begin{tikzpicture}[
node distance=6mm,
block/.style={
draw,
rounded corners=2pt,
minimum width=3.2cm,
minimum height=0.75cm,
align=center,
font=\footnotesize
},
arrow/.style={
-{Latex[length=2.5mm]},
thick
}
]

\node[block] (model) {Camera};

\node[block, below=of model] (fk) {Object Detection};

\node[block, below=of fk] (ik) {Target Position \\ (X, Y, z)};

\node[block, below=of ik] (angles) {Inverse Kinematics};

\node[block, below=of angles] (JA) {Joint Angles \\$(\theta_1,\ldots,\theta_6)$};

\node[block, below=of JA] (serial) {Serial Communication\\(via MATLAB)};

\node[block, below=of serial] (stm) {STM32H743\\Motion Controller};

\node[block, below=of stm] (stepper) {Interrupt-Driven\\Step Generation};

\node[block, below=of stepper] (robot) {6-DOF\\Robotic Manipulator};

\draw[arrow] (model) -- (fk);
\draw[arrow] (fk) -- (ik);
\draw[arrow] (ik) -- (angles);
\draw[arrow] (angles) -- (JA);
\draw[arrow] (JA) -- (serial);
\draw[arrow] (serial) -- (stm);
\draw[arrow] (stm) -- (stepper);
\draw[arrow] (stepper) -- (robot);

\end{tikzpicture}

\caption{\footnotesize Kinematic modelling and motion execution workflow. The manipulator is modelled using a MATLAB \texttt{rigidBodyTree}, inverse kinematics generates the required joint angles, and the computed commands are transmitted to the STM32H743 controller for synchronized stepper motor actuation.}
\label{fig:kinematics_workflow}
\end{figure}
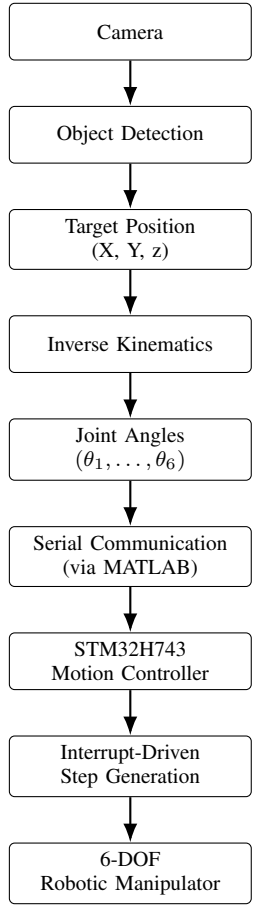

The complete modelling and motion-execution workflow is summarised in
Fig.~\ref{fig:kinematics_workflow}. The joint angles computed by the inverse kinematics solver are transmitted from
MATLAB to the STM32H743-based controller via a serial communication interface.
The embedded firmware converts the received joint positions into synchronised
step commands for the six stepper motors using the interrupt-driven pulse
generation algorithm described in Section~\ref{sec:firmware}.

To maintain continuous and smooth motion, long trajectories are divided into a
sequence of smaller motion segments before transmission\cite{Craig}. This chunked-motion
strategy prevents pauses caused by communication latency between the host
computer and the embedded controller while ensuring coordinated motion of all
six joints throughout the trajectory.

\section{Control Interface}
\begin{figure}[t]
  \centering
  \includegraphics[width=\columnwidth]{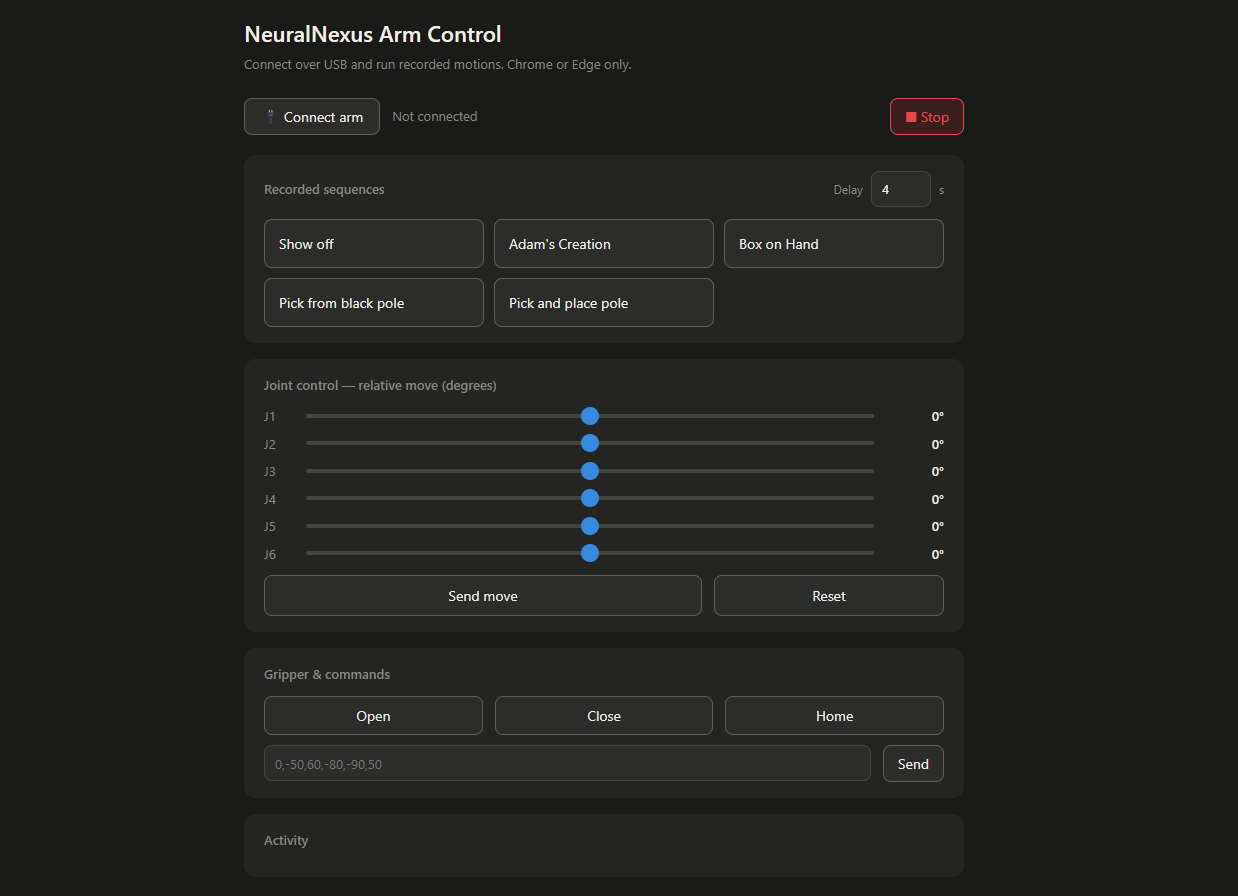}
  \caption{Browser-native control panel (Web Serial API).}
  \label{fig:ui}
\end{figure}

The arm is controlled from a browser using the Web Serial API: any Chromium-based browser connects directly to the controller's serial port, with no vendor IDE or native driver required on the host. The panel (Fig. ~\ref{fig:ui}) provides per-joint sliders for manual jogging, a set of pre-recorded motion sequences, and gripper commands \textit{(G,1 to open; G,0 to close)}, together with a Stop control. Cartesian moves currently follow a separate path: joint angles are computed by the MATLAB solver of Section V-B and streamed to the controller over the same serial link. Porting the solver into the browser is left for future work.

\section{Vision Pipeline}
\label{sec:vision_pipeline}

We are using a Raspberry Pi 5, 4 GB RAM device with a Pi Camera 3 for the vision-related implementation. The camera detects the targeted object and converts it into a bounding box, and the centre pixel is transformed into real-world coordinates. These coordinates are then used in inverse kinematics to move the arm to the relevant position.

One of the main design compromises we made is that we used a Pi Camera 3 module instead of a depth camera. The Pi Camera 3 module can only detect the $x$ and $y$ coordinates in a given frame. Therefore, our main idea was to have a fixed robotic arm position with a known height (in our case, $z = 41$ cm). The arm always comes to the fixed position, takes the pixel value, transforms it into $x$ and $y$ coordinates, and the $z$ coordinate is already known. Therefore, all $x$, $y$, and $z$ coordinates are known, and these coordinates are sent to the inverse kinematics section.

Target points are considered valid only when they lie within the calibrated workspace plane
\cite{hartley2004,opencv_homography}.

The complete vision pipeline includes image capturing, camera calibration and image undistortion, colour-based object segmentation, image-centre estimation, planar coordinate transformation, and the generation of robot target coordinates.

\begin{equation}
\begin{split}
\text{Camera Image}
&\rightarrow \text{Undistortion}
\rightarrow \text{HSV Segmentation} \\
&\rightarrow \text{Object Centroid}
\rightarrow \text{Homography} \\
&\rightarrow (X,Y,Z)
\rightarrow \text{Inverse Kinematics}.
\end{split}
\end{equation}

\subsection{Camera Calibration and Image Undistortion}

Geometric camera calibration was performed before object localisation in
order to determine the camera intrinsic parameters and compensate for lens
distortion. Planar chessboard patterns are widely used for this purpose since
known points on the calibration target can be associated with their observed
image coordinates \cite{zhang2000,opencv_calibration}.

\begin{figure}[t]
    \centering

    \includegraphics[width=0.31\columnwidth]
    {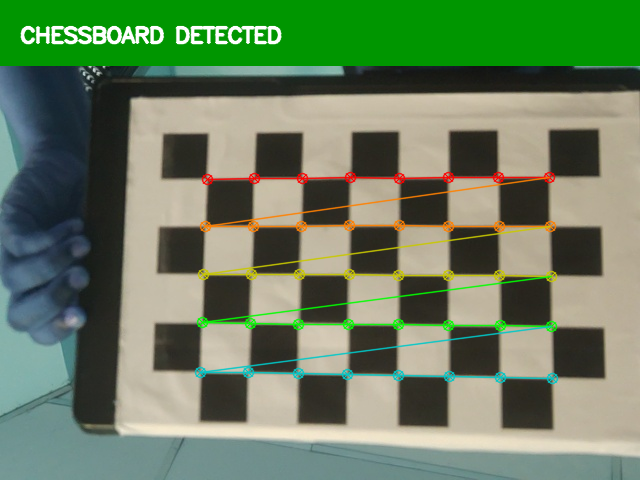}
    \hfill
    \includegraphics[width=0.31\columnwidth]
    {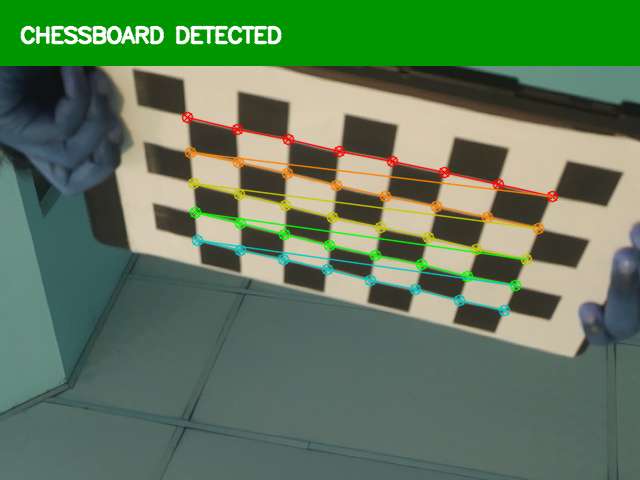}
    \hfill
    \includegraphics[width=0.31\columnwidth]
    {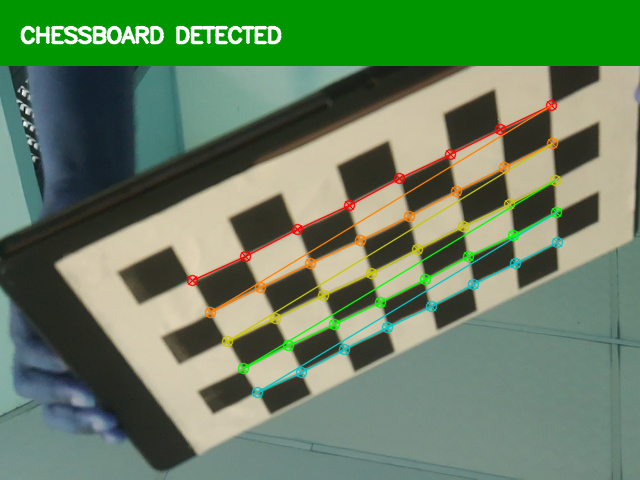}

    \vspace{1mm}

    \includegraphics[width=0.31\columnwidth]
    {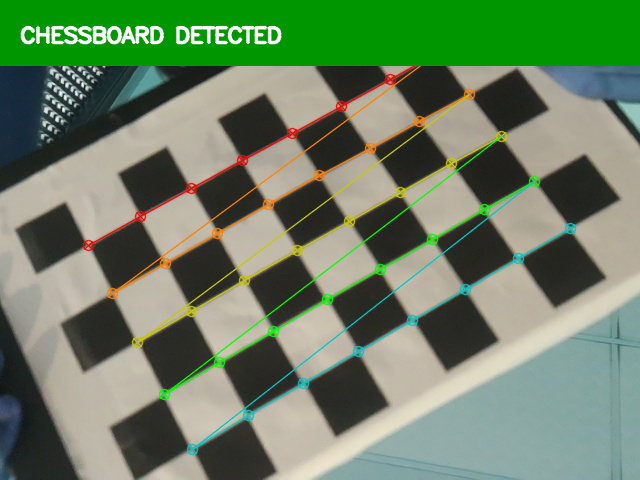}
    \hfill
    \includegraphics[width=0.31\columnwidth]
    {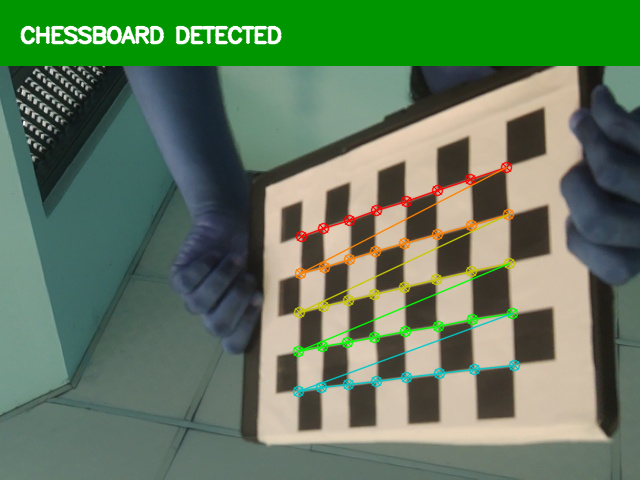}
    \hfill
    \includegraphics[width=0.31\columnwidth]
    {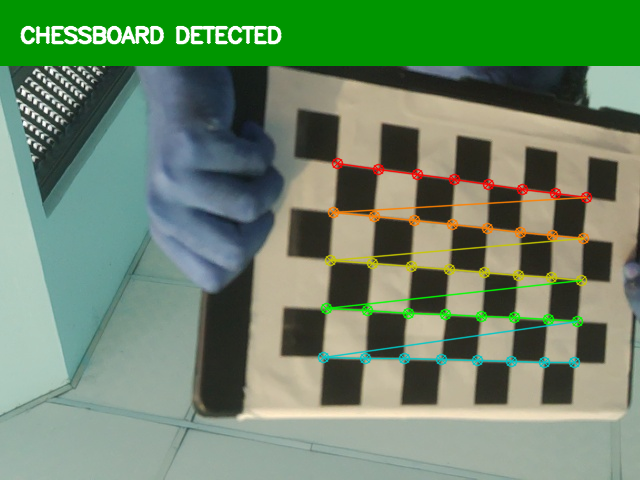}

    \vspace{1mm}

    \includegraphics[width=0.31\columnwidth]
    {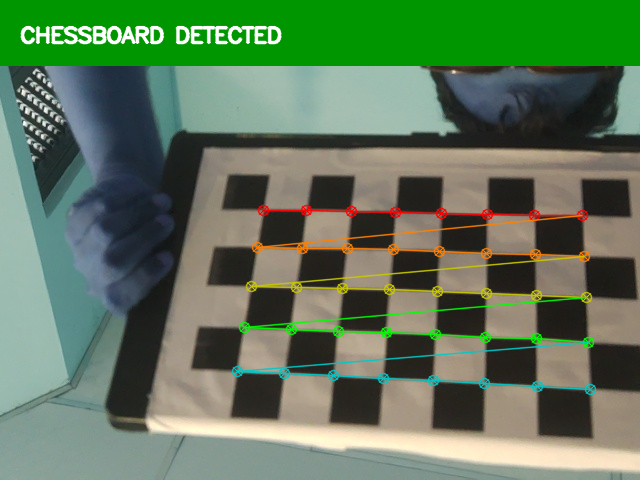}
    \hfill
    \includegraphics[width=0.31\columnwidth]
    {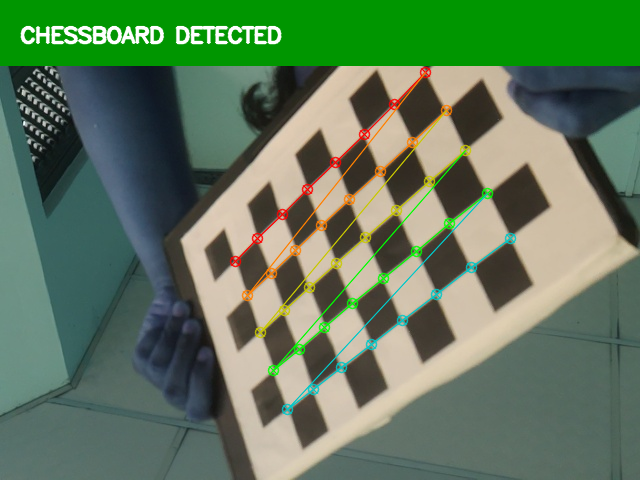}
    \hfill
    \includegraphics[width=0.31\columnwidth]
    {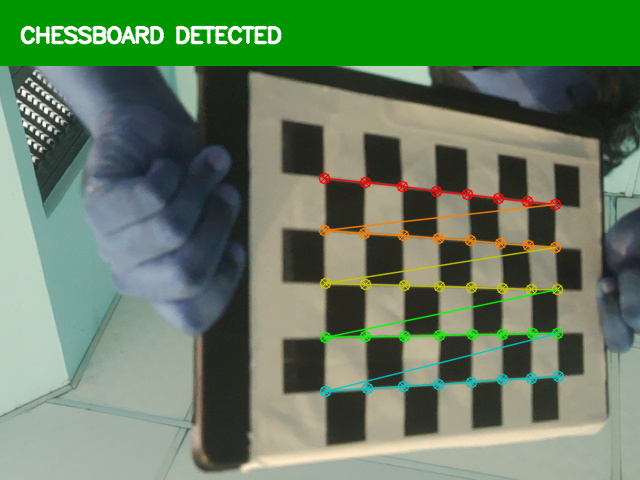}

    \caption{Detected $8 \times 5$ inner-corner patterns from nine
    calibration images captured at different positions and orientations.}
    \label{fig:chessboard_grid}
\end{figure}

We are using a chessboard with
$32~\mathrm{mm}$ squares for the calibration. In our case, 20 different chessboard images were used to estimate the intrinsic camera matrix and lens-distortion parameters. By facing the chessboard at different angles and different distances from the camera, these parameters were estimated; Fig.~\ref{fig:chessboard_grid} shows the detected corner pattern in nine of the twenty calibration images. One important point is not to move the Pi Camera during the calibration. The only movable object should be the printed chessboard, and we need to make sure that the printed chessboard is completely flat without any curves.

The intrinsic camera matrix is represented as

\begin{equation}
\mathbf{K} =
\begin{bmatrix}
f_x & 0 & c_x \\
0 & f_y & c_y \\
0 & 0 & 1
\end{bmatrix},
\label{eq:general_camera_matrix}
\end{equation}

where $f_x$ and $f_y$ denote the focal lengths expressed in pixels, and
$(c_x,c_y)$ denotes the principal point \cite{hartley2004}.

The calibration procedure produced the following intrinsic camera matrix:

\begin{equation}
\mathbf{K} =
\begin{bmatrix}
917.9335 & 0 & 329.7344 \\
0 & 917.0094 & 241.1553 \\
0 & 0 & 1
\end{bmatrix}.
\label{eq:camera_matrix}
\end{equation}

Thus,

\begin{equation}
f_x = 917.9335,
\qquad
f_y = 917.0094,
\end{equation}

and

\begin{equation}
(c_x,c_y)=(329.7344,\;241.1553).
\end{equation}

In addition to intrinsic calibration, lens distortion was estimated. The
standard OpenCV distortion model includes three radial coefficients and two
tangential coefficients \cite{opencv_calibration}:

\begin{equation}
\mathbf{D}
=
\begin{bmatrix}
k_1 & k_2 & p_1 & p_2 & k_3
\end{bmatrix}.
\end{equation}

The experimentally obtained distortion parameters were

\begin{equation}
\mathbf{D}
=
\begin{bmatrix}
0.03355 &
-0.65939 &
-0.00434 &
0.00502 &
4.40902
\end{bmatrix}.
\label{eq:distortion_coefficients}
\end{equation}

Here, $k_1$, $k_2$, and $k_3$ correspond to radial distortion, while $p_1$
and $p_2$ correspond to tangential distortion. Radial distortion primarily
causes displacement that increases with distance from the optical centre,
whereas tangential distortion is associated with imperfect alignment of the
lens and imaging plane \cite{opencv_calibration}.

The estimated matrices were stored as
\texttt{camera\_matrix.npy} and \texttt{dist\_coeffs.npy}. During operation,
each captured frame was undistorted using these calibration parameters before
object localisation.

\subsection{Colour-Based Target Segmentation}

Initial object-detection experiments considered a YOLO-based detector.
For the initial object-detection testing cases, we used \texttt{yolo26n.pt}, \texttt{yolo11n.pt}, and custom-trained YOLO detection models that primarily detect cardboard boxes.

However, in the final stages, a colour-based segmentation method was selected because it has a higher FPS than the YOLO models, as it does not require processing through a heavy YOLO model. In addition, using this method makes object detection highly usable, as it can detect any object of any shape in the given colour. Therefore, there is no need to train the model separately for each object.

In the pipeline, we first converted the BGR representation used by
OpenCV into the HSV colour space. HSV is useful for colour-based segmentation because the hue component provides a direct representation of colour type, while saturation and value describe colour purity and brightness,
respectively \cite{opencv_inrange}.

The red target is extracted by applying HSV range thresholding. Because red appears around the boundary of the cyclic hue coordinate, two hue intervals
can be combined to form the final red-object mask.

HSV colour is represented using $H$ (hue), $S$ (saturation), and $V$ (brightness).
In OpenCV, the hue range is defined cyclically from 0 to 179. Therefore, the hue value for red is between

\[
0 \leq H \leq 10
\qquad \text{and} \qquad
170 \leq H \leq 179.
\]

Therefore, for the final red mask, we need to combine these two ranges.

\begin{equation}
M_{\mathrm{red}}
=
M_1 \lor M_2,
\label{eq:redmask}
\end{equation}

where $M_1$ and $M_2$ denote the binary masks obtained from the two selected
red HSV ranges. OpenCV's range-thresholding operation provides a direct
implementation of this type of HSV segmentation \cite{opencv_inrange}.

\subsection{Contour Detection and Centroid Estimation}

\begin{figure}[t]
    \centering

    \includegraphics[width=0.48\columnwidth]
    {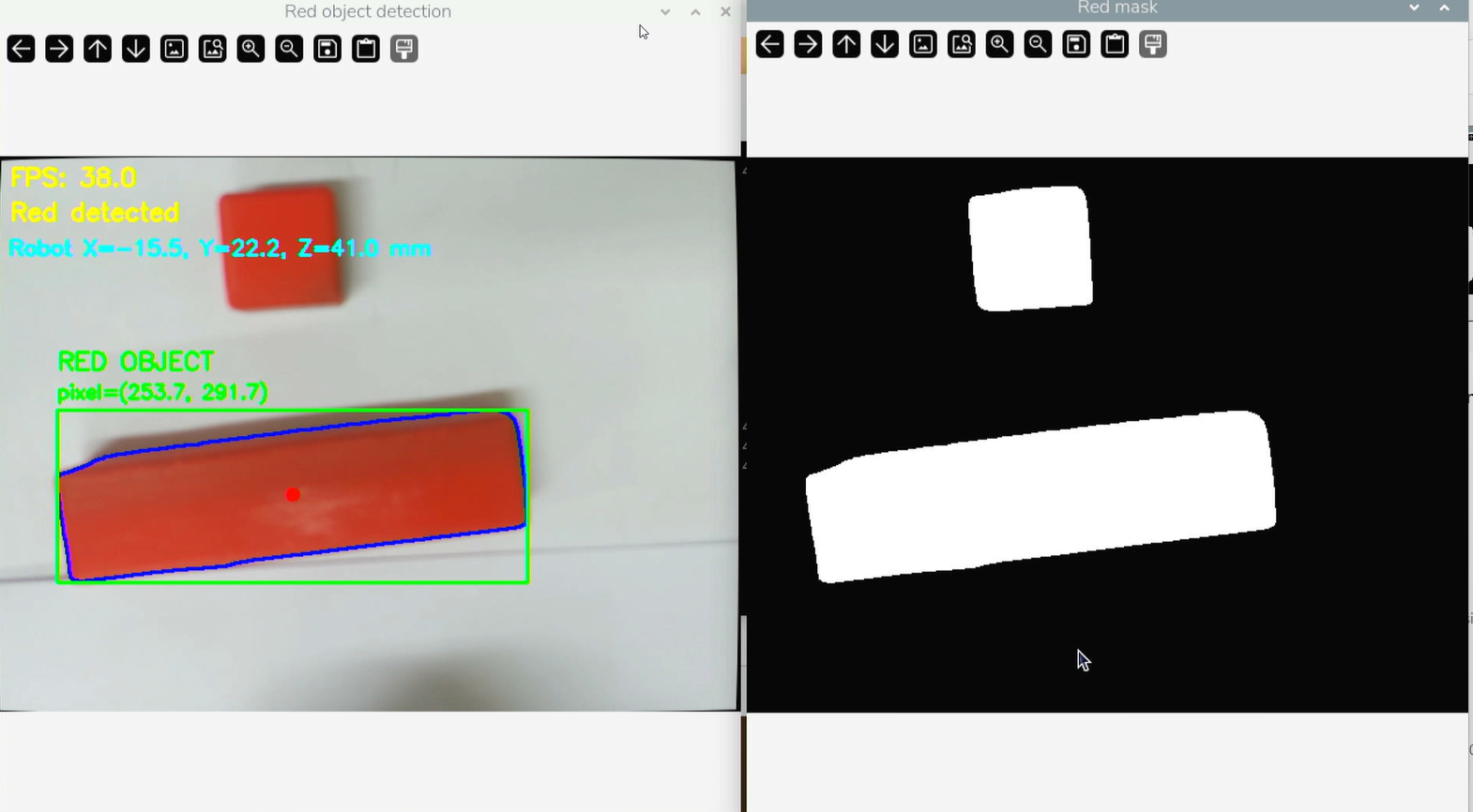}
    \hfill
    \includegraphics[width=0.48\columnwidth]
    {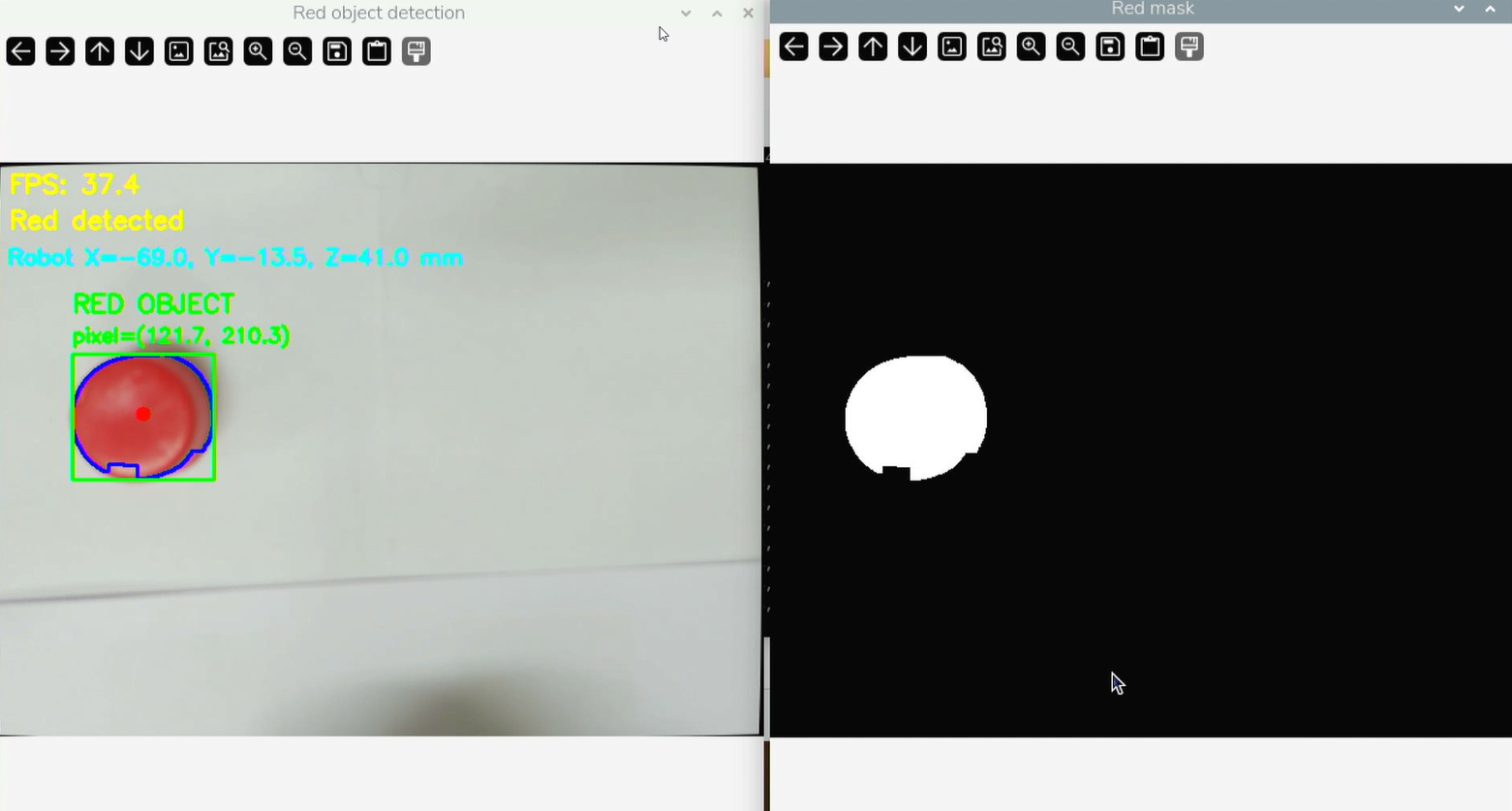}

    \vspace{1mm}

    \includegraphics[width=0.48\columnwidth]
    {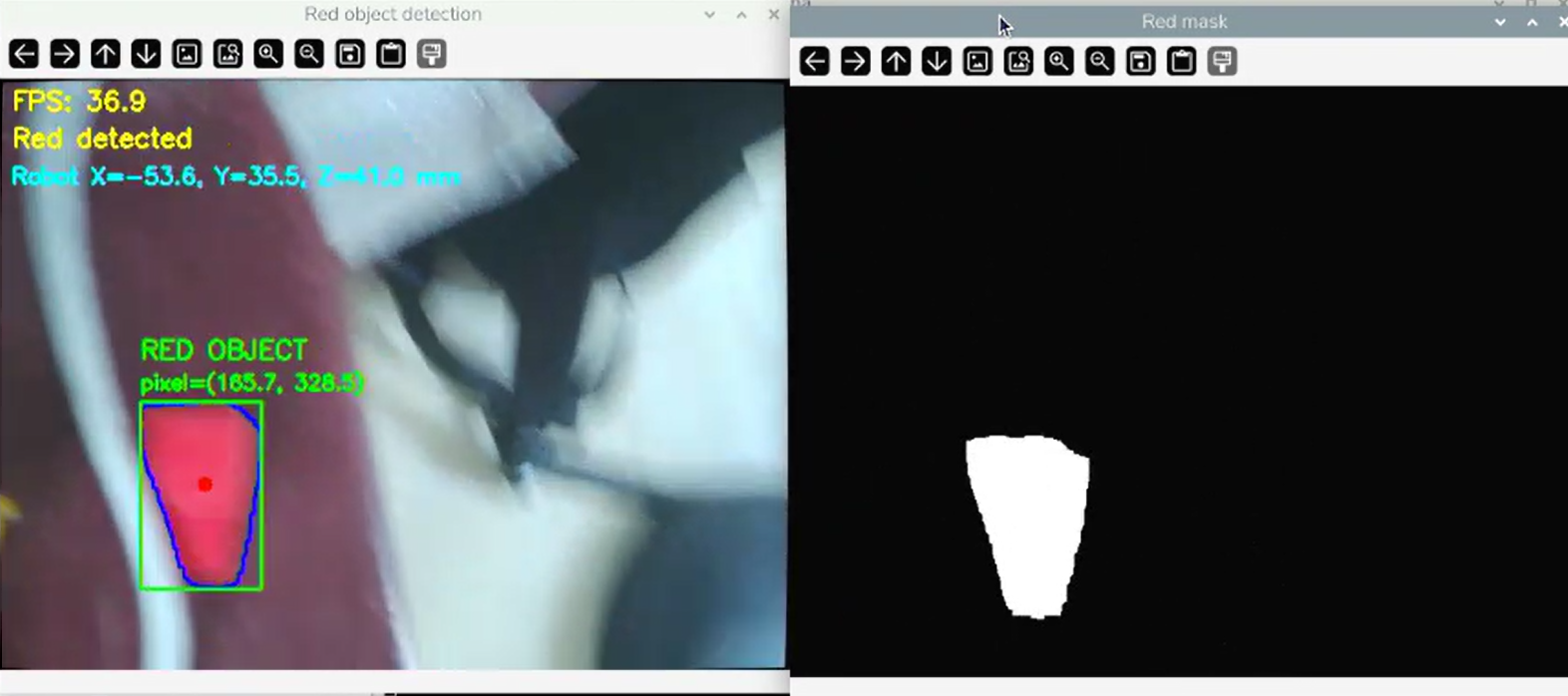}
    \hfill
    \includegraphics[width=0.48\columnwidth]
    {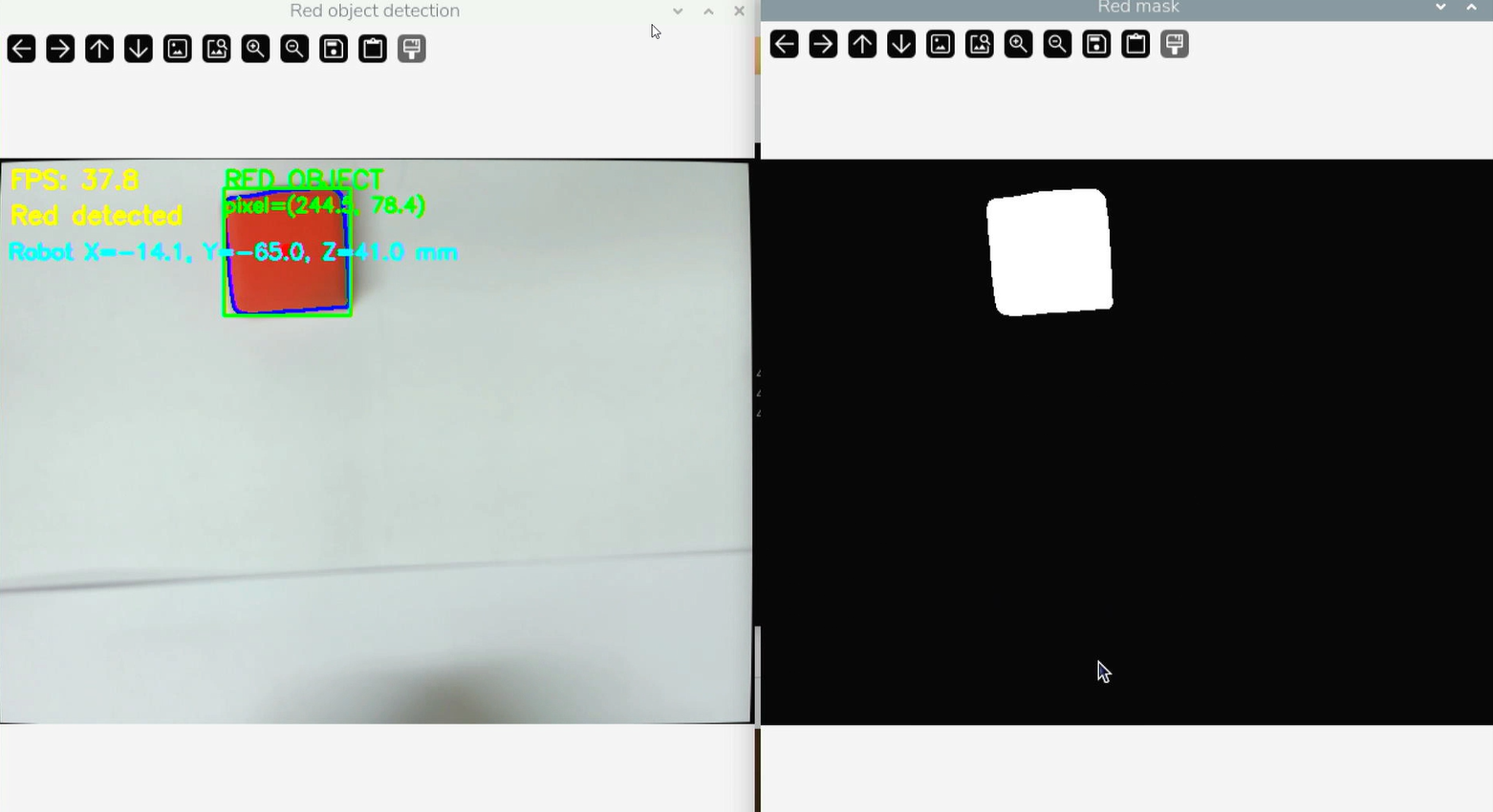}

    \caption{Detected object-processing stages used for target localisation.}
    \label{fig:vision_detection_stages}
\end{figure}

Fig.~\ref{fig:vision_detection_stages} shows representative targets at each
stage of this process. Contours are extracted from the segmented binary image to identify connected
regions corresponding to candidate objects. Small contours caused by image
noise or unwanted red regions are rejected based on their area.

Let the detected contours be

\begin{equation}
\mathcal{C}
=
\{C_1,C_2,\ldots,C_N\}.
\end{equation}

The contour area is calculated for each region, and the valid target contour
can be selected according to the expected target size.

The centre of the selected target is determined using spatial image moments.
Image moments provide quantities including region area and centroid
\cite{opencv_contours}. For a binary region, the spatial moments may be
written as

\begin{equation}
M_{pq}
=
\sum_u \sum_v u^p v^q I(u,v).
\end{equation}

The centroid coordinates are then obtained from

\begin{equation}
u_c =
\frac{M_{10}}{M_{00}},
\end{equation}

\begin{equation}
v_c =
\frac{M_{01}}{M_{00}}.
\end{equation}

Therefore, the detected object location in the image plane is

\begin{equation}
\mathbf{p}_{\mathrm{image}}
=
\begin{bmatrix}
u_c\\
v_c
\end{bmatrix}.
\end{equation}

These coordinates are expressed in pixels and therefore require transformation
into the physical coordinate system of the robotic workspace.

\subsection{Planar Workspace Calibration}

We use a planar homography to map the pixel values to real-world $(x,y)$ coordinates. Homography calibration uses a $3 \times 3$ matrix to perform the projective transformation between the two planes
\cite{hartley2004,opencv_homography}.

During workspace calibration, we select known physical locations. The known physical coordinate

\begin{equation}
\mathbf{P}_i=(X_i,Y_i)
\end{equation}

is associated with the corresponding image coordinate

\begin{equation}
\mathbf{p}_i=(u_i,v_i).
\end{equation}

\begin{figure}[t]
    \centering
    \includegraphics[
        width=\columnwidth
    ]{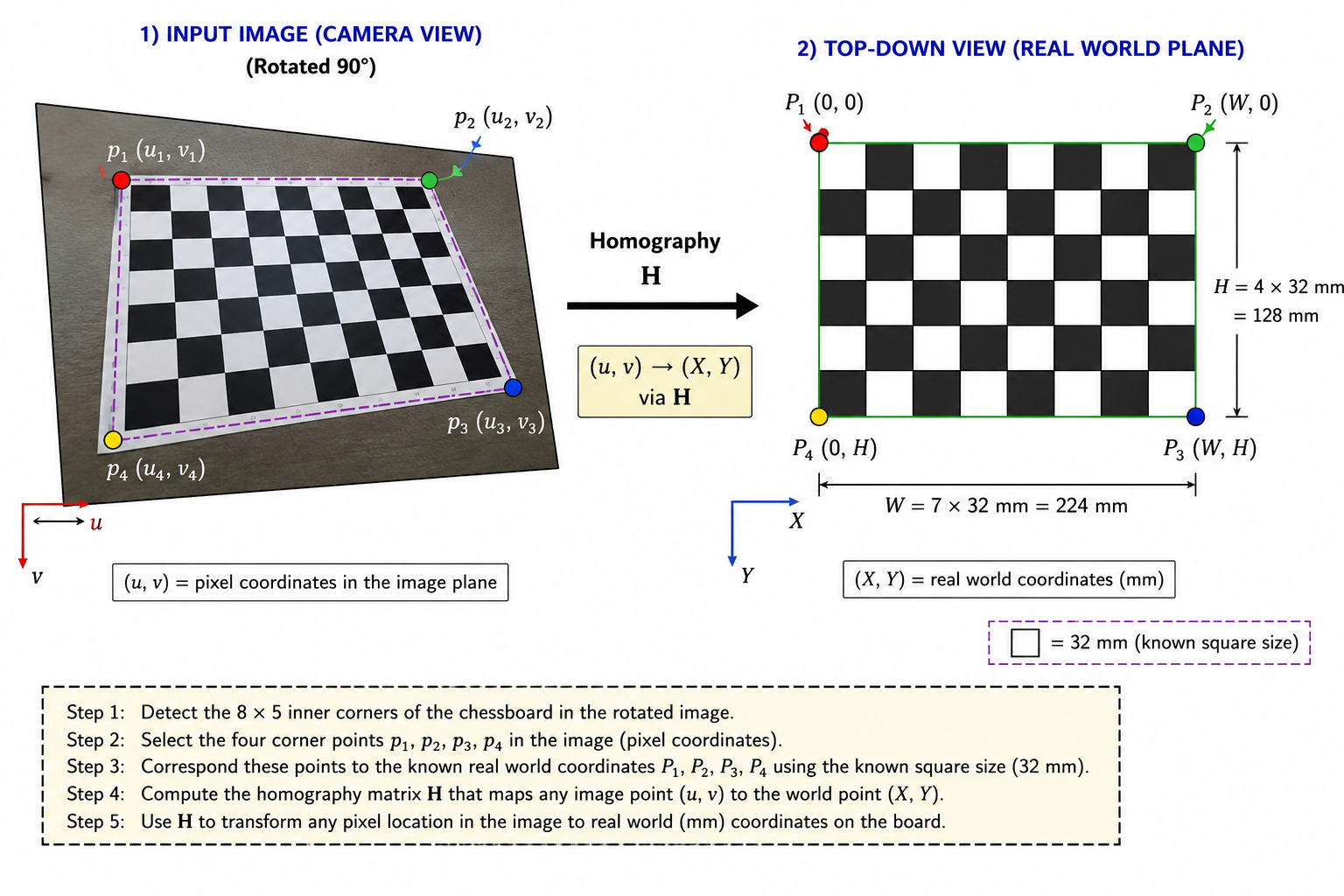}
    \caption{Visualisation of planar homography calibration using the
    $32~\mathrm{mm}$ chessboard. The detected image-plane corner points
    are mapped to a top-down representation of the calibrated workspace.}
    \label{fig:homography_visualisation}
\end{figure}

Using these point correspondences, the homography matrix

\begin{equation}
\mathbf{H}
=
\begin{bmatrix}
h_{11} & h_{12} & h_{13}\\
h_{21} & h_{22} & h_{23}\\
h_{31} & h_{32} & h_{33}
\end{bmatrix}
\end{equation}

was estimated, as illustrated in Fig.~\ref{fig:homography_visualisation}. For planar scenes, image-to-plane mappings of this form can be
estimated directly from corresponding points \cite{opencv_homography}.

For the target centroid $(u_c,v_c)$,

\begin{equation}
\begin{bmatrix}
X'\\
Y'\\
W'
\end{bmatrix}
=
\mathbf{H}
\begin{bmatrix}
u_c\\
v_c\\
1
\end{bmatrix}.
\label{eq:homography_mapping}
\end{equation}

The corresponding physical coordinates are obtained through homogeneous
normalisation,

\begin{equation}
X=\frac{X'}{W'},
\qquad
Y=\frac{Y'}{W'}.
\end{equation}

The calibrated transformation matrix was stored as
\texttt{homography\_matrix.npy} and reused during real-time operation.
Because both the camera and workspace remain fixed, the same transformation
remains applicable provided that their relative geometry does not change.

\subsection{Generation of Robot Target Coordinates}

The homography transformation provides the planar location

\begin{equation}
\mathbf{p}_{xy}
=
\begin{bmatrix}
X\\
Y
\end{bmatrix}.
\end{equation}

Since the target object is placed on the known workspace plane, its vertical
coordinate is defined from the corresponding workspace height,

\begin{equation}
Z=Z_{\mathrm{table}}.
\end{equation}

The final target coordinate supplied to the robot controller is therefore

\begin{equation}
\mathbf{p}_{\mathrm{target}}
=
\begin{bmatrix}
X\\
Y\\
Z_{\mathrm{table}}
\end{bmatrix}.
\end{equation}

The resulting Cartesian position is then provided to the
inverse-kinematics algorithm of the 6-DOF robotic arm to determine the joint
configuration necessary to approach the detected object.

Therefore, the complete transformation can be summarised as

\begin{equation}
\boxed{
(u_c,v_c)
\rightarrow
(X,Y)
\rightarrow
(X,Y,Z_{\mathrm{table}})
\rightarrow
\boldsymbol{\theta}
}
\end{equation}

where $\boldsymbol{\theta}$ represents the set of robot joint variables.

\section{Bill of Materials}
\label{sec:bom}

Table~\ref{tab:bom} summarises the cost of every component and
fabrication service by subsystem; the full itemised breakdown appears in
Appendix~\ref{app:bom}. The total of \$1{,}512.33 supersedes the
approximate figure given in Table~\ref{tab:comparison}.

\begin{table}[t]
\caption{Bill of materials summary by subsystem. Full itemised breakdown in Appendix~\ref{app:bom}.}
\label{tab:bom}
\centering
\renewcommand{\arraystretch}{1.15}
\begin{tabular}{@{}lr@{}}
\toprule
\textbf{Subsystem} & \textbf{Cost (\$)} \\
\midrule
Actuation --- motors, drivers, gripper & \$300.38 \\
Control, PCB \& sensing electronics & \$252.36 \\
Vision compute & \$178.80 \\
Mechanical parts (gears, pulleys, belts, bearings) & \$296.82 \\
Local hardware --- misc.\ components, fasteners & \$149.96 \\
Enclosure / structural fabrication & \$293.08 \\
\midrule
BOM subtotal (parts + fabrication) & \$1471.42 \\
Import / PCB duties & \$40.92 \\
\midrule
\textbf{Total} & \textbf{\$1512.33} \\
\bottomrule
\end{tabular}
\end{table}

\section{Engineering Methodology and Lessons Learned}
The platform began on a development board and was later ported to the custom PCB.
That migration exposed a cascade of issues whose root causes are worth recording,
since they are the kind of problem a reproducing team will also encounter.

\subsection{Phantom Encoder Wander}
A stationary joint appeared to report an encoder spread of roughly
$\pm 72^\circ$. The cause was not signal integrity: error sentinel values of
$-1$ were being included in a running average of encoder readings, producing an
apparent wander. The decisive clue was that \emph{slowing} the SPI clock made the
effect \emph{worse} rather than better, which ruled out analogue/timing causes
and pointed at the data path. Excluding the sentinels from the average removed
the artefact.

\subsection{CubeMX Regeneration Trap}
Regenerating the project from CubeMX re-introduced \texttt{MX\_SDMMC1\_SD\_Init()}
and \texttt{MX\_FATFS\_Init()}, both of which trap in \texttt{Error\_Handler()}
when no SD card is present. The immediate workaround is to comment these out after
each regeneration; the durable fix is to disable SDMMC1 and FATFS in the
\texttt{.ioc} so they are not regenerated.

\section{Results and Validation}
\begin{figure}[t]
  \centering
  \includegraphics[width=\columnwidth]{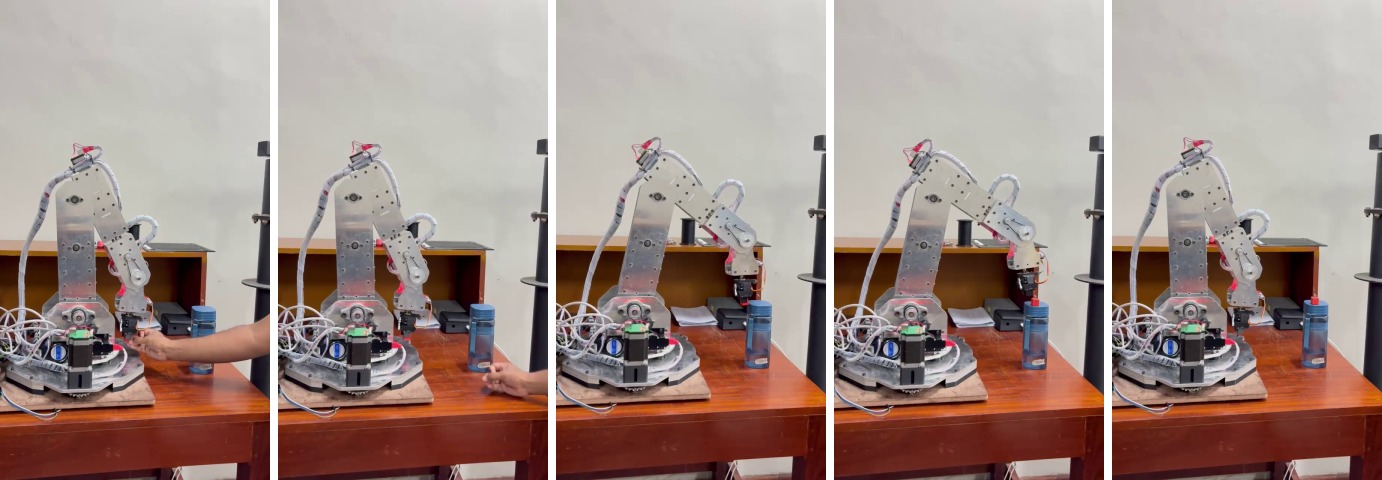}
  \includegraphics[width=\columnwidth]{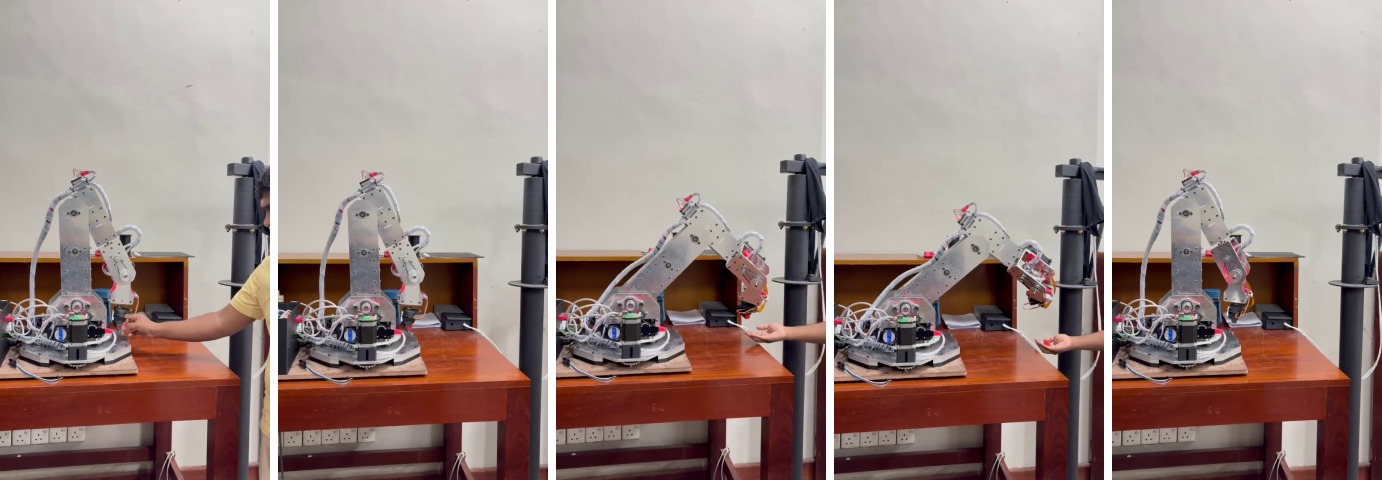}
  \includegraphics[width=\columnwidth]{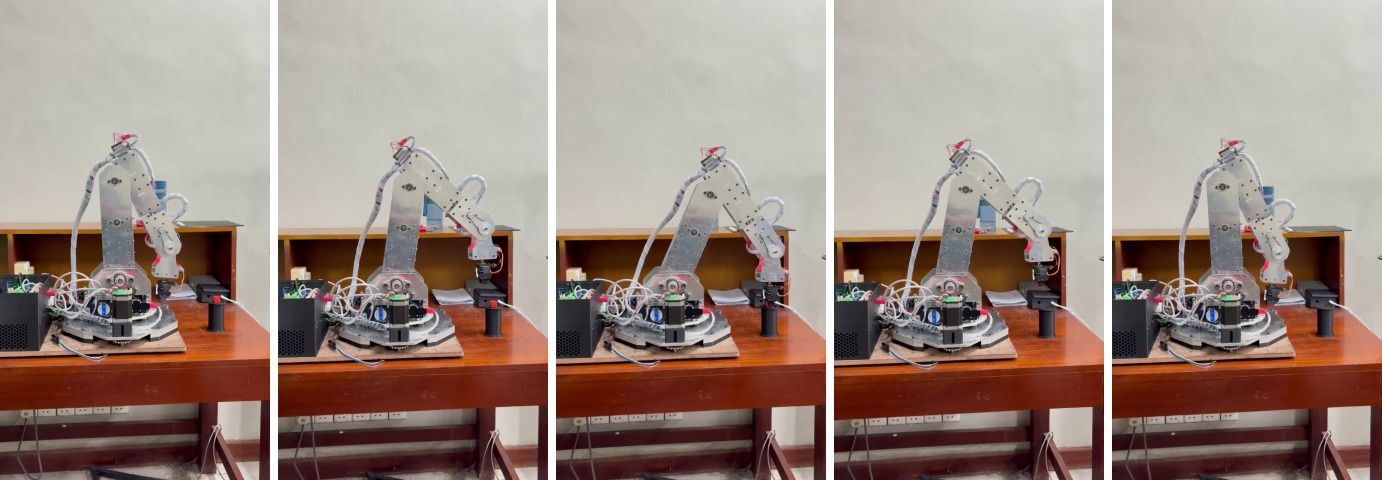}
  
  \caption{Selected frames from Video 2, showing a full
pick-and-place sequence: \videotwourl.}
  \label{fig:frames}
\end{figure}

The arm actuates on all six axes and executes both manual jog commands and the
pre-recorded sequences from the browser interface. Motion timing is consistent
with the step-rate model above (approximately 4\,s for a 60$^\circ$ move at
400\,steps/s). Table~\ref{tab:performance} summarises the measured performance of the
platform. Selected frames are shown in Fig.~\ref{fig:frames}; a full
demonstration is provided in Video 2.\footnote{Video 2 (pick-and-place sequence): \videotwourl}

\begin{table}[htpb]
\caption{Experimental Performance Summary}
\label{tab:performance}
\centering
\renewcommand{\arraystretch}{1.15}
\begin{tabular}{lc}
\hline
\textbf{Metric} & \textbf{Value} \\
\hline
Positioning repeatability & $\pm 1.3$ mm \\
Pick-and-place cycle time & $12.5$ s \\
Task success rate &  $97\%$ \\
Maximum payload & $2000$ g \\
Maximum joint speed & $10^\circ$/s \\
\hline
\end{tabular}
\end{table}

\section{Future Work}

Future work will investigate the use of the NeuralNexus Arm as a
reconfigurable platform for different robotic manufacturing and automation
applications. Potential applications include robotic additive manufacturing
and multi-axis 3D printing, where an extrusion-based end effector can be
mounted to the wrist for freeform material deposition. The manipulator may
also be adapted for robotic laser engraving and low-power laser cutting,
where the six-degree-of-freedom architecture enables controlled tool
orientation along complex paths.

Other possible applications include automated adhesive and sealant
dispensing, component sorting, surface
inspection, and laboratory automation. Interchangeable end effectors could
further allow the same platform to support light-duty machining, polishing,
painting, and similar robotic process-automation tasks.

\section{Conclusion}
The NeuralNexus Arm demonstrates that a capable, reproducible 6-DOF stepper arm
can be built and controlled by a small team using a single microcontroller, a
pragmatic mix of onboard and external drivers, and a browser-native control
interface with no proprietary host software. By documenting not only the finished
design but also the bring-up failures and their root causes, we aim to make the
platform genuinely reproducible.

\section*{Availability}
The complete design is released openly. The main repository, including CAD,
PCB design files and documentation are available at \repotagurl, and the firmware codebase at \firmwarerepourl. Three supplementary videos are available: Video 1, an overview of the
assembled arm (\armvideourl); Video 2, a full
pick-and-place sequence (\videotwourl, shown in Fig.~\ref{fig:frames}); and Video 3, pick-and-place demonstration 2 (\videothreeurl).

\section*{Acknowledgment}

The authors would like to thank the Department of Electronic and Telecommunication Engineering, University of Moratuwa, for providing the facilities and support required for this work. We also extend our gratitude to our advisors and colleagues for their valuable guidance, feedback, and assistance throughout the development of the NeuralNexus Arm.

\appendix[Detailed Bill of Materials]
\label{app:bom}

\begin{table*}[p]
\centering
\caption{Detailed bill of materials. Costs converted from LKR at 1~LKR = \$0.00298 (Aug.\ 2026).}
\label{tab:bom-full}
\small
\setlength{\tabcolsep}{4pt}
\renewcommand{\arraystretch}{1.1}
\begin{tabular}{@{}p{5.6cm}cccll@{}}
\toprule
\textbf{Item} & \textbf{Qty} & \textbf{Unit (\$)} & \textbf{Total (\$)} & \textbf{Supplier} & \textbf{Used in} \\
\midrule
\multicolumn{6}{l}{\textit{Actuation --- motors, drivers, gripper}} \\
NEMA 23 stepper motor (Local) & 1 & \$22.65 & \$22.65 & Local Shop & J1 (base) \\
NEMA 17 stepper motor (Local) & 3 & \$7.45 & \$22.35 & AliExpress & Wrist joints \\
NEMA 24 stepper motors CL57T drivers+ gearboxes + PSU (StepperOnline) & 2 & \$119.20 & \$238.40 & Stepper Online & J2, J3 \\
DM542 stepper driver (Leadshine, DM542) & 1 & \$9.83 & \$9.83 & Local Shop & J1 \\
Gripper / end-effector hardware (3D Printed) & --- & --- & \$7.15 & Local Shop & Gripper \\
\multicolumn{3}{r}{\textit{Subtotal}} & \textit{\$300.38} & & \\
\addlinespace
\multicolumn{6}{l}{\textit{Control, PCB \& sensing electronics}} \\
PCB components (bulk order) & --- & --- & \$108.77 & LCSC & Controller PCB \\
PCB fabrication (4-layer) & 1 & \$85.90 & \$85.90 & JLCPCB & Controller PCB \\
Encoder PCB fabrication & --- & --- & \$21.34 & JLC PCB & 6x AS5047P encoder boards \\
USB extender cable & 1 & \$2.38 & \$2.38 & Local & Controller / host link \\
Network cable & 1 & --- & \$14.75 & Local & Vision (Pi) \\
HDMI cable & 1 & --- & \$11.92 & Local & Pi display \\
Camera cable & 1 & --- & \$7.30 & Raspberry Pi & Vision (Pi Camera) \\
\multicolumn{3}{r}{\textit{Subtotal}} & \textit{\$252.36} & & \\
\addlinespace
\multicolumn{6}{l}{\textit{Vision compute}} \\
Raspberry Pi 5 (Raspberry Pi Foundation, 4 GB RAM) & 1 & \$134.10 & \$134.10 & Raspberry Pi & Vision pipeline \\
Pi Camera 3 & 1 & \$44.70 & \$44.70 & Raspberry Pi & Vision pipeline \\
\multicolumn{3}{r}{\textit{Subtotal}} & \textit{\$178.80} & & \\
\addlinespace
\multicolumn{6}{l}{\textit{Mechanical parts (gears, pulleys, belts, bearings)}} \\
Gears / pulleys / belts (order 1) & --- & --- & \$47.24 & AliExpress & Drivetrain \\
Gears / pulleys / belts (order 2) & --- & --- & \$77.73 & AliExpress & Drivetrain \\
Gears / pulleys / belts (order 3) & --- & --- & \$144.72 & AliExpress & Drivetrain \\
Bearings (reorder) & --- & --- & \$11.28 & AliExpress & Joints \\
Bearings (reorder) & --- & --- & \$15.85 & Local & Joints \\
\multicolumn{3}{r}{\textit{Subtotal}} & \textit{\$296.82} & & \\
\addlinespace
\multicolumn{6}{l}{\textit{Local hardware --- misc.\ components, fasteners}} \\
Misc components (power supply, fans, cables) & --- & --- & \$63.98 & Local & Electronic Box \\
TMC2209 + other items & --- & --- & \$21.16 & BigTreeTech & NEMA 17 \\
Metal rods (16mm, 8mm, 5mm) & --- & --- & \$12.96 & Local & Shafts \\
Electronic components & --- & --- & \$16.33 & Local & Various \\
Generic hardware & --- & --- & \$3.61 & Local & Various \\
Nails / fasteners & --- & --- & \$1.19 & Local & Enclosure \\
Bearings and screws & --- & --- & \$30.74 & Local & Various \\
\multicolumn{3}{r}{\textit{Subtotal}} & \textit{\$149.96} & & \\
\addlinespace
\multicolumn{6}{l}{\textit{Enclosure / structural fabrication}} \\
Plywood & --- & --- & \$4.77 & Local & Enclosure \\
Laser cutting (3 orders) & --- & --- & \$123.37 & Local & Frame / enclosure panels \\
Lathe and CNC Services & --- & --- & \$10.10 & Local & Frame \\
Metal Rod Cutting and CNC & --- & --- & \$4.47 & Local & Shafts \\
Frame hardware & --- & --- & \$1.73 & Local & Enclosure \\
3D printing (Distribution Box, Electronic Enclosure) (PLA \& ABS) & --- & --- & \$59.24 & Local & Custom mounts / brackets \\
Aluminium sheet, 4mm (6061) & 1 & \$89.40 & \$89.40 & Local & Links / frame \\
\multicolumn{3}{r}{\textit{Subtotal}} & \textit{\$293.08} & & \\
\addlinespace
\midrule
\multicolumn{3}{r}{\textbf{BOM subtotal (parts + fabrication)}} & \textbf{\$1471.42} & & \\
\multicolumn{3}{r}{Import / PCB duties} & \$40.92 & & \\
\midrule
\multicolumn{3}{r}{\textbf{Total}} & \textbf{\$1512.33} & & \\
\bottomrule
\end{tabular}
\end{table*}

\clearpage

\bibliographystyle{IEEEtran}
\bibliography{references}

\end{document}